\documentclass[journal]{IEEEtran}
\ifCLASSINFOpdf
  \usepackage[pdftex]{graphicx}
\else
\fi
\usepackage{amsmath}
\usepackage{amssymb}
\usepackage{mathtools}

\usepackage{graphicx}
\usepackage{multirow}
\usepackage{subcaption}
\usepackage{url}
\usepackage{color}
\usepackage{hyperref}

\newcommand{\reffig}[1]{Figure~\ref{#1}}
\newcommand{\reftab}[1]{Table~\ref{#1}}
\newcommand{\refsec}[1]{Section~\ref{#1}}
\renewcommand{\theenumi}{\roman{enumi}}

\newcommand*{\defeq}{\mathrel{\vcenter{\baselineskip0.5ex \lineskiplimit0pt
                     \hbox{\scriptsize.}\hbox{\scriptsize.}}}%
                     =}

\usepackage[acronym]{glossaries}

\newacronym{vio}{VIO}{Visual-Inertial Odometry}
\newacronym{vo}{VO}{Visual Odometry}
\newacronym{slam}{SLAM}{Simultaneous Localization And Mapping}
\newacronym{uwb}{UWB}{Ultra-Wideband}
\newglossaryentry{kf}{
	name={KF},
	description={Keyframe},
	first={\glsentrydesc{kf} (\glsentrytext{kf})},
  	plural={KFs},
  	descriptionplural={Keyframes},
  	firstplural={\glsentrydescplural{kf} (\glsentryplural{kf})}
}
\newglossaryentry{mp}{
	name={MP},
	description={Map Point},
	first={\glsentrydesc{mp} (\glsentrytext{mp})},
  	plural={MPs},
  	descriptionplural={Map Points},
  	firstplural={\glsentrydescplural{mp} (\glsentryplural{mp})}
}
\newglossaryentry{uav}{
	name={UAV},
	description={Unmanned Aerial Vehicle},
	first={\glsentrydesc{uav} (\glsentrytext{uav})},
  	plural={UAVs},
  	descriptionplural={Unmanned Aerial Vehicles},
  	firstplural={\glsentrydescplural{uav} (\glsentryplural{uav})}
}
\newacronym{dof}{DoF}{Degree-of-freedom}
\newglossaryentry{imu}{
	name={IMU},
	description={Inertial Measurement Unit},
	first={\glsentrydesc{imu} (\glsentrytext{imu})},
  	plural={IMUs},
  	descriptionplural={Inertial Measurement Units},
  	firstplural={\glsentrydescplural{imu} (\glsentryplural{imu})}
}
\newacronym{ekf}{EKF}{Extended Kalman Filter}
\newacronym{admm}{ADMM}{Alternating Direction Method of Multipliers}
\newglossaryentry{rmse}{
	name={RMSE},
	description={Root Mean Square Error},
	first={\glsentrydesc{rmse} (\glsentrytext{rmse})},
  	plural={RMSEs},
  	descriptionplural={Root Mean Square Errors},
  	firstplural={\glsentrydescplural{rmse} (\glsentryplural{rmse})}
}

\begin{document}
%
\title{Distributed Variable-Baseline Stereo SLAM from two UAVs}
%
%
%

\author{Marco~Karrer and
        Margarita~Chli \\
Vision for Robotics Lab, ETH Zurich, Switzerland\\
}
\maketitle

\begin{abstract}
\gls{vio} has been widely used and researched to control and aid the automation of navigation of robots especially in the absence of absolute position measurements, such as GPS.
However, when observable landmarks in the scene lie far away from the robot's sensor suite, as it is the case at high altitude flights, the fidelity of estimates and the observability of the metric scale degrades greatly for these methods.
Aiming to tackle this issue, in this article, we employ two \glspl{uav} equipped with one monocular camera and one \gls{imu} each, to exploit their view overlap and relative distance measurements between them using \gls{uwb} modules onboard to enable collaborative \gls{vio}.
In particular, we propose a novel, distributed fusion scheme enabling the formation of a virtual stereo camera rig with adjustable baseline from the two \glspl{uav}.
In order to control the \gls{uav} agents autonomously, we propose a decentralized collaborative estimation scheme, where each agent hold its own local map, achieving an average pose estimation latency of 11ms, while ensuring consistency of the agents' estimates via consensus based optimization.
Following a thorough evaluation on photorealistic simulations, we demonstrate the effectiveness of the approach at high altitude flights of up to 160m, going significantly beyond the capabilities of state-of-the-art \gls{vio} methods.
%

%
%
Finally, we show the advantage of actively adjusting the baseline on-the-fly over a fixed, target baseline, reducing the error in our experiments by a factor of two.
\end{abstract}



\textbf{Video -- \url{https://youtu.be/SdL4Jb-BQ28}}

\section{Introduction}\label{sec:introduction}
Awareness of a robot's pose in previously unseen environments is one of the key elements towards enabling autonomous navigation of robots.
\begin{figure}[t]
    \centering
    \vspace{2mm}
    ~ 
        \includegraphics[trim={0 0 0 0}, clip, width=1.0\columnwidth]{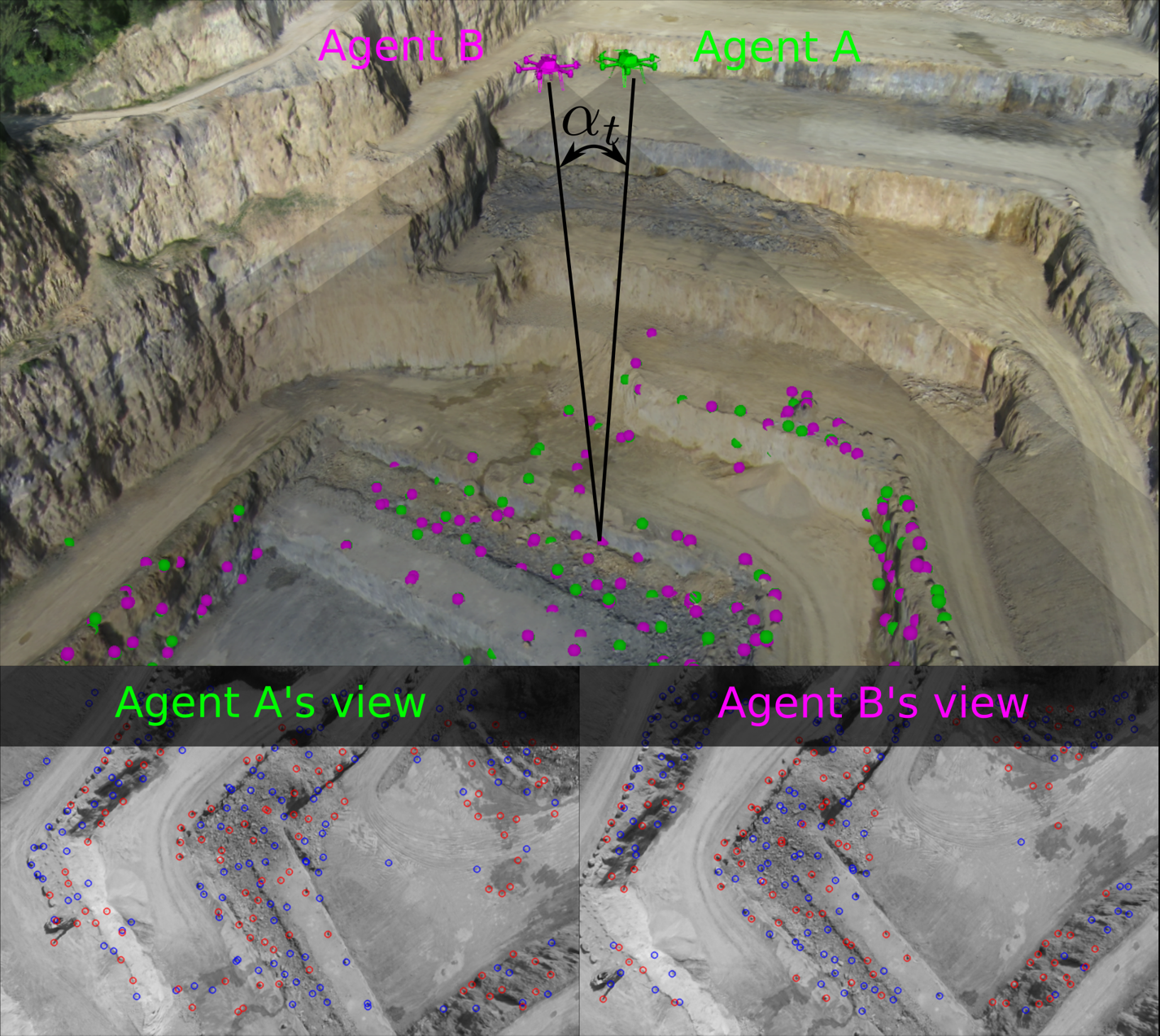}
    \caption{Using two UAVs, equipped with one IMU and one monocular camera each, while measuring the relative distance between them using an ultrawideband module, the proposed sensor fusion framework is able to reliably estimate both UAVs' poses in metric scale at the particularly challenging case of high-altitude fights. 
 This is achieved by adjusting the baseline between the UAVs adaptively, to realize a desired triangulation angle $\alpha_{t}$. 
 In the distributed implementation, each agent holds its own version of the map, indicated by the green and magenta points in the top image, enabling low latency pose estimation of the UAVs, while consistency across the agents is achieved via a consensus based optimization scheme.
 The red and blue circles in the bottom two images indicate keypoints with and without 3D point associated with them, respectively.}
\label{fig:title_image}
\end{figure}
While GPS is of tremendous help towards achieving this goal, its availability and quality are still limited, e.g. indoors or close to structure, hence in a general scenario strong reliance on GPS can be a critical limitation. 
Addressing this issue, research into \gls{slam} has received quite some attention as it potentially enables ego-motion estimation only using the sensors carried onboard the robot. \\
Due to the low cost and low power consumption of cameras and their ability to capture rich information about the environment, they have become a popular sensor choice for performing \gls{slam}.
With the first systems \cite{Davison:etal:PAMI2007}, \cite{Eade:Drummond:CVPR2006}, \cite{Klein:Murray:ISMAR2007} demonstrating the ability to estimate the egomotion up to scale using only a single camera, soon more mature and complete systems, such as \cite{Mur-Artel:etal:TRO2015}, emerged.
While the scale ambiguity inherently present in monocular systems can be resolved, for example, by using a stereo camera rig \cite{Mur-Artel:etal:TRO2017}, the lack of information in-between frames renders purely vision-based systems sensitive to fast motions, which are rather common when employed onboard small \glspl{uav}.
Therefore, the complementary characteristics of \glspl{imu} and cameras have been explored extensively to develop robust and accurate state estimation by the means of \gls{vio} systems, such as \cite{Bloesch:etal:IJRR2017}, \cite{Leutenegger:etal:IJRR2015}, \cite{Qin:etal:TRO2018}, which quickly became a standard for autonomous navigation of \glspl{uav} \cite{Oleynikova:etal:JFR2019}. \\
While \gls{vio} techniques have been successfully applied in situations where the scene is in the close proximity to the camera (e.g. in autonomous driving, in low-altitude flights or operations indoors), in scenarios where the scene is further away from the camera, such as at high altitude flights, \gls{vio} methods experience difficulties to establish similar levels of accuracy.
%
%
As all \gls{vio} methods rely on triangulation of scene landmarks, the scene depth estimates become more uncertain the smaller the triangulation angle gets.
For monocular \gls{vio}, this angle is determined by the camera motion and the scene depth.
As the overall scale of monocular \gls{vio} conceptually, is determined via the integration of \gls{imu} measurements, with increasing scene depth, a larger range of motion is required in order to obtain a large enough triangulation angle.
This leads to the integration of more \gls{imu} readings, which at some point results in a numerically weakly constrained scale. 
%
The same effect can be observed when using stereo cameras, where the fixed baseline dictates the maximum depth that can be reliably detected \cite{Hinzmann:etal:IROS2019} and thus, limiting the effectiveness of the imposed constraints on the scale.
%
The choice of the stereo baseline is, thus, a crucial parameter and ideally, one would be able to modify this parameter on-the-fly depending on scene depth or accuracy requirements of the task at hand \cite{Gallup:etal:CVPR2008}. \\
Inspired by the idea of using two \glspl{uav} as agents equipped with one camera each, to form a virtual stereo camera rig \cite{Achtelik:etal:IROS2011}, we propose a novel, complete system utilizing view-overlap across two agents together with relative distance measurements obtained via \gls{uwb} in order to collaboratively estimate the pose of these two cameras.
As the two agents are capable of modifying their relative poses, the baseline between their cameras can be adjusted accordingly, e.g. achieving a desired triangulation angle.
In order to allow for a low latency pose estimation onboard the agents, independently of network delays, we propose a novel decentralized optimization architecture allowing each agent to hold their own estimate of the map, while ensuring consistency of common scene estimates amongst them. 
In brief, the main contributions of this work are the following:
\begin{itemize}
\item a novel, real-time sensor fusion framework combining monocular visual and inertial data from two \glspl{uav}, and relative distance measurements between them (e.g. using ultra-wideband modules on each agent) enabling a reliable relative pose estimation of each other even at high altitudes,
\item the adaptation of the framework of \cite{Peng:etal:SIAM2016} for asynchronous multi-agent estimation to enable sliding-window bundle adjustment in a decentralized fashion for the first time to the best of our knowledge,
\item a thorough evaluation of the proposed system using photorealistic simulation showing the improvements over state-of-the-art stereo-inertial methods at higher altitudes, and
\item demonstration of the advantage of an adjustable baseline in the proposed two-UAV stereo setup for accurate pose estimation by the means of a simple high-level formation controller.
\end{itemize}

\section{Related Work}\label{sec:related_work}
Alongside the emergence of vision based \gls{slam} for single robots \cite{Bloesch:etal:IJRR2017, Klein:Murray:ISMAR2007,Leutenegger:etal:IJRR2015, Mur-Artel:etal:TRO2017, Qin:etal:TRO2018}, research into multi-robot systems started to attract attention recently.
The collaborative framework proposed in \cite{Forster:etal:IROS2013}, demonstrates how \gls{kf} data from  cameras mounted on different \glspl{uav} can be combined into a global map.
Designed for mapping visual-inertial data from multiple agents, \cite{Karrer:etal:ICRA2018_global} builds up a global map, showcasing the benefit of collaboration on the overall trajectory error.
Utilizing Manhattan-World like structures where available in order to constrain the map optimization, \cite{Guo:etal:ICRA2016} propose a large  scale visual-inertial mapping framework and \cite{Mohanarajah:etal:TASE2015} utilize a cloud computer to generate 3D maps, collaboratively.
%
While these approaches are focused on mapping, other systems aim to distribute the processes of the \gls{slam} pipeline between the agents and a server \cite{Riazuelo:etal:RAS2014, Schmuck:Chli:JFR2018, Karrer:etal:RAL2018}, promising to reduce the computational load onboard the agents and to make map data generated from one agent available to all the robotic team.
On the other hand, there also exist multi-session frameworks, such as \cite{Schneider:etal:RAL2018, Qin:etal:TRO2018} that allow the user to re-localize in previously generated maps.
In order to allow these systems to scale up to larger teams of robots, research effort has been aiming to avoid a central server entity and perform all operations in a distributed fashion instead.
Many works focus on optimizing a specific aspect of the multi-agent system for a distributed setup, such as place recognition \cite{Cieslewski:Scaramuzza:RAL2017}, robustness \cite{Zhang:etal:IJARS2018}, or efficient data exchange \cite{Cieslewski:etal:ICRA2018}. \\
All the aforementioned frameworks make use of collaboration of some form ranging from combining map data into a larger map, to re-using parts of a map created from other agents. 
Nonetheless, the tracking front-ends of these systems do not require any tight collaboration amongst them and therefore, share the same limitations as their single-agent counterparts, for example, degrading quality of the pose estimate at higher altitudes.
A counter example is the collaborative system CoSLAM \cite{Zou:Tan:PAMI2013}, which addresses the problem of inaccuracies due to dynamic objects in the scene by using multiple freely moving cameras with view-overlap.
This approach allows reliable pose estimation of cameras, which only observes the dynamic scene points in collaboration with their neighboring cameras. 
However, the system requires all image data to be collected at a single access point and makes heavy use of GPU computation, limiting the applicability of the approach to a robotic problem.\\
The problem of degrading quality of \gls{vio} state estimation at high altitudes has recently received some attention in the literature.
In wind turbine inspection using \glspl{uav}, for example, \cite{Teixeira:etal:RAL2018} proposed a framework with two \glspl{uav}, one equipped with LED markers and a GPS module and the other one with one camera that can detect the markers and one camera for turbine inspection. 
For the estimation, the \gls{uav} with the cameras observes the other \gls{uav}'s markers and flies close to the turbine, where the GPS signal is disturbed, while the \gls{uav} with the markers flies further away from the turbine to secure more reliable GPS reception, while staying in the field of view of the observing \gls{uav}.
While this method fits well to inspection of structures, where GPS reception is reliable at least slightly further away from the structure, it is clear that unreliable or imprecise GPS readings have a strong effect on the quality of the pose estimates.
Other methods deal with creating a large stereo baseline by placing the two cameras at the tips of a fixed-wing aircraft \cite{Hinzmann:etal:IROS2019, Hinzmann:etal:ICRA2018}, where the authors model and correct for movements between the cameras due to deformation of the structure essentially continuously estimating the stereo extrinsics.
In the work of \cite{Achtelik:etal:IROS2011} and \cite{Karrer:etal:ICRA2018} the estimation of the relative transformation between two independently moving \glspl{uav} equipped with monocular cameras by the means of (relative) motion and view overlap is investigated.
While conceptually, such a configuration can be interpreted as a stereo camera, the absolute scale of the baseline only gets estimated via motion (e.g. \gls{imu} measurements), hence suffering from the same limitations as monocular \gls{vio}.
Nonetheless, taking inspiration from \cite{Achtelik:etal:IROS2011}, here we propose a framework performing \gls{vio} with two agents, each equipped with a monucular camera and an \gls{imu} in tight collaboration with each other
%
using relative distance measurements between the agents, e.g. using \gls{uwb} modules, the scale ambiguity of the variable baseline setup can be addressed, allowing to effectively obtain a virtual stereo camera with a baseline that can be adjusted according to the scene and accuracy requirements.
In contrast to other collaborative estimators with tight collaboration on the front-end such as \cite{Zou:Tan:PAMI2013}, we propose a decentralized architecture allowing for low-latency pose estimation independent of communication delays and fitting well within the bandwidth limitations of a standard WiFi module.

\section{Preliminaries}\label{sec:preliminaries}
\subsection{Notation}
\label{ssec:notation}
Throughout this work, we use small bold letters (e.g. $\mathbf{a}$) to denote vector values, capital bold letters (e.g. $\mathbf{A}$) to denote matrices and plain capitals (e.g. $A$) denote coordinate frames. 
To indicate a submatrix formed by the rows $r_{i}$ to $r_{j}$ and the columns $c_{l}$ to $c_{k}$ of $\mathbf{A}$ we use the notation $\mathbf{A}\left[r_{i},r_{j}; c_{l}, c_{k}\right]$, where a $1$ based indexing is used.
A vector $\mathbf{x}$ expressed in the coordinate frame $A$, is denoted as $_{A}\mathbf{x}$. 
Rigid body transformations from coordinate frame $B$ to coordinate frame $A$ are denoted by $\mathbf{T}_{AB}$, comprising the translational part of the transformation $\mathbf{p}_{AB}$ and the rotational part $\mathbf{R}_{AB}$. 
For notational brevity, at times we use quaternions $\mathbf{q}_{AB}$ interchangeably with such rotation matrices. 
The concatenation of two quaternions $\mathbf{q}_{1}$ and $\mathbf{q}_{2}$ is denoted by $\mathbf{q}_{1} \circ \mathbf{q}_{2}$, whereas the rotation of vector $\mathbf{v}$ by a quaternion is denoted by $\mathbf{q}(\mathbf{v})$.
Values that correspond to a prediction are indicated with $\hat{\cdot}$, whereas measurements are denoted with $\tilde{\cdot}$.
Finally, sets of variables are denoted using capital calligraphic letters (e.g. $\mathcal{A}$).

\subsection{Asynchronous-parallel ADMM}
\label{ssec:async_admm}
The \gls{admm} was first introduced by \cite{Magnus:Hestenes:JOTA} and is an algorithm that aims to solve problems of the following form:
\begin{equation}\label{eq:admm_base_problem}
\begin{aligned}
\underset{\mathbf{x}, \mathbf{y}}{\mathrm{minimize}
}& \quad f(\mathbf{x}) + g(\mathbf{y}) \\
\mathrm{subject \: to}&\quad  \mathbf{A}\mathbf{x} + \mathbf{B} \mathbf{y} = \mathbf{w},
\end{aligned}
\end{equation}
for proper, convex functions $f(\cdot)$ and $g(\cdot)$, 
by forming the augmented Lagrangian $L_{\gamma}$ introducing the dual variables $\mathbf{z}$ and the penalty weight $\gamma$ as follows:
\begin{equation}\label{eq:augmented_lagrangian}
\resizebox{.99\hsize}{!}{$
L_{\gamma}(\mathbf{x}, \mathbf{y}, \mathbf{z})=f(\mathbf{x}) +g(\mathbf{y})+ \mathbf{z}^{T} (\mathbf{A} \mathbf{x} + \mathbf{B}\mathbf{y} - \mathbf{w}) + \frac{\gamma}{2} \lVert \mathbf{A}\mathbf{x} + \mathbf{B}\mathbf{y} - \mathbf{w}\rVert_{2}^{2} ~.
$} 
\end{equation}
The \gls{admm} algorithm solves the problem in Eq. \eqref{eq:augmented_lagrangian} by performing the following update steps iteratively:
\begin{eqnarray}
\mathbf{x}^{k+1} =& \! &\underset{\mathbf{x}}{\mathrm{arg \, min} \:} L_{\gamma}(\mathbf{x}, \mathbf{y}^{k}, \mathbf{z}^{k}) \label{eq:admm_base_it1} \\
\mathbf{y}^{k+1} =& \! &\underset{\mathbf{y}}{\mathrm{arg \, min} \:} L_{\gamma}(\mathbf{x}^{k+1}, \mathbf{y}, \mathbf{z}^{k})\label{eq:admm_base_it2} \\
\mathbf{z}^{k+1} =& \! &\mathbf{z}^{k} + \gamma (\mathbf{A}\mathbf{x}^{k+1} + \mathbf{B}\mathbf{y}^{k+1} - \mathbf{w}) \label{eq:admm_base_it3} ~,
\end{eqnarray}
where $k$ is the iteration number.
While the problem formulation in Eq. \eqref{eq:admm_base_problem} corresponds to generic constraint optimization, unconstrained optimization problems with additive cost functions can be brought into an equivalent form, allowing for a distributed implementation over $N$ nodes as used by \cite{Zhang:etal:PAMI2018}, for example. 
The form of this distributed problem is given by
\begin{equation}\label{eq:admm_consensus_problem}
\begin{aligned}
\underset{\mathbf{x_{i}}}{\mathrm{mininize}}& \quad \sum\limits_{i = 1}^{N} f_{i}(\mathbf{x}_{i}) \\
\mathrm{subject \: to }&\quad \mathbf{x}_{i} = \mathbf{S}_{i} \mathbf{y}, \quad i = 1,\ldots, N ~,
\end{aligned}
\end{equation}
where $\mathbf{S}_{i}$ corresponds to an indicator matrix, selecting the entries in $\mathbf{y}$ corresponding to the variables of $\mathbf{x}_{i}$. 
Note that in relation to Eq. \eqref{eq:admm_base_problem}, here $g(\mathbf{y}) = 0$, $\mathbf{w} =\mathbf{0}$, $\mathbf{A}$ is the identity matrix and $\mathbf{B}$ corresponds to the stacked version of the indicator matrices $-\mathbf{S}_{i}$.
As shown in \cite{Zhang:etal:PAMI2018}, this formulation reduces the update step of $\mathbf{y}$ in Eq. \eqref{eq:admm_base_it2} to a simple averaging operation (consensus). \\
While the algorithm proposed by \cite{Zhang:etal:PAMI2018} can parallelize the most expensive operation of solving the local optimization problems in Eq. \eqref{eq:admm_base_it1}-\eqref{eq:admm_base_it3}, in order to compute the consensus terms, all data needs to pass through one central point. 
So in effect, the synchronized structure of the algorithm dictates the frequency of the \gls{admm} iterations to be equal to the slowest participating node. 
The algorithmic framework ``ARock" introduced in \cite{Peng:etal:SIAM2016}, specifically addresses this limitation allowing to arrive to an asynchronous implementation of the \gls{admm} algorithm, which we adopt in this work. 
A brief overview of the ARock \gls{admm} algorithm presented in \cite{Peng:etal:SIAM2016} is provided below, while the specific details of the algorithm related to the proposed framework are introduced in \refsec{ssec:optimization}.\\
In the general setup, consider a set of nodes $1,\ldots,N$, forming a graph connecting the nodes by a set of edges $\mathcal{E} = \lbrace (i, j) | \text{if node }i \text{ connects to node } j, i < j  \rbrace$. 
A node can be seen as a computational unit holding a partial, local estimate $\mathbf{x}_{i}$ of the optimization state variable $\mathbf{x}$.
%
For the sake of notational simplicity, in the following, all local variables $\mathbf{x}_{i}$ are assumed to be realizations of the full state $\mathbf{x}$, i.e. the dimensions of all $\mathbf{x}_{i}$ are equal to the dimension of $\mathbf{x}$.
Additionally, we assume that not every node can communicate with every other node and the edges in $\mathcal{E}$ is given by pairwise set of nodes that can communicate with each other. 
%
%
As shown in \refsec{ssec:optimization}, this can be generalized to sharing only a subset of variables between the nodes. 
In such a setup, the problem formulated in Eq. \eqref{eq:admm_consensus_problem} can be expressed as:
\begin{equation}\label{eq:arock_admm_base}
\begin{aligned}
\underset{\mathbf{x}_{i}, \mathbf{y}_{ij}}{\mathrm{minimize}\:} &\sum\limits_{i = 1}^{N} f_{i}(\mathbf{x}_{i}) \\
\mathrm{subject\, to} \quad&\mathbf{x}_{i} = \mathbf{y}_{ij}, \: \mathbf{x}_{j} = \mathbf{y}_{ij} \quad \forall (i,j) \in \mathcal{E} ~. 
\end{aligned}
\end{equation}
Let $\mathcal{E}_{i}$ denote the set of edges connected to node $i$ and let $|\mathcal{E}_{i}|$ denote its cardinality. 
Furthermore, let $\mathcal{L}_{i} = \lbrace j | (j,i)\in \mathcal{E}_{i}, j < i\rbrace$ and $\mathcal{R}_{i} = \lbrace j | (i,j)\in \mathcal{E}_{i}, j > i \rbrace$, separating the ordered indices to the left and the right of $i$, respectively. 
For every pair of constraints $\mathbf{x}_{i}=\mathbf{y}_{ij}$ and $\mathbf{x}_{j} =\mathbf{y}_{ij}$, such that $(i,j) \in \mathcal{E}$ in Eq. \eqref{eq:arock_admm_base}, the dual variables $\mathbf{z}_{ij,i}$ and $\mathbf{z}_{ij,j}$ get associated on node $i$ and $j$, respectively.
Consequently, the \gls{admm} iterations for every node $i$ can be written as
%
\begin{align}
\mathbf{x}^{k+1}_{i} &= \underset{\mathbf{x}_{i}}{\mathrm{arg\, min}} \: f_{i}(\mathbf{x}_{i}) + (\mathbf{z}_{i}^{k})^{T}\mathbf{x}_{i} + \frac{\gamma}{2}|\mathcal{E}_{i}| \cdot \lVert \mathbf{x}_{i} \rVert^{2} \label{eq:arock_min} \\
\mathbf{z}_{li,i}^{k+1} &= \mathbf{z}_{li,i}^{k} - \eta_{k} \cdot \left((\mathbf{z}_{li,i}^{k} + \bar{\mathbf{z}}_{li,l})/ 2 +\gamma \mathbf{x}_{i}^{k+1} \right), \quad \forall l \in \mathcal{L}_{i}  \label{eq:arock_left} \\
\mathbf{z}_{ir,i}^{k+1} &= \mathbf{z}_{ir,i}^{k} - \eta_{k} \cdot \left((\mathbf{z}_{ir,i}^{k} + \bar{\mathbf{z}}_{ir,r})/ 2 + \gamma \mathbf{x}_{i}^{k+1} \right), \quad \forall r \in \mathcal{R}_{i} \label{eq:arock_right} ~,
\end{align}
where the variables denoted with $\bar{\cdot}$ represent the latest received values from the neighboring nodes and the weight $\eta_{k}$ is a factor to account for communication delays and the corresponding potential use of outdated dual variables. 
Note that $\mathbf{z}_{i}^{k}$ corresponds to 
\begin{equation} \label{eq:sum_of_duals}
\mathbf{z}_{i}^{k} = \sum\limits_{l \in \mathcal{L}_{i}} \bar{\mathbf{z}}_{li,l}^{k} + \sum\limits_{r \in \mathcal{R}_{i}} \bar{\mathbf{z}}_{ir,r}^{k} ~.
\end{equation}
Finally, after every iteration of Eq. \eqref{eq:arock_min}-\eqref{eq:arock_right} for node $i$, the updated dual variables $\mathbf{z}_{ji,i}^{k+1}$ get communicated to the nodes $j \in \mathcal{E}_{i}$ connected to $i$.

\subsection{Z-Spline based 6DoF Pose Interpolation}
\label{ssec:z_splines}
\begin{figure}[t]
    \centering
    \vspace{2mm}
    ~ 
    \begin{subfigure}[b]{0.48\columnwidth}
        \includegraphics[width=1.0\columnwidth]{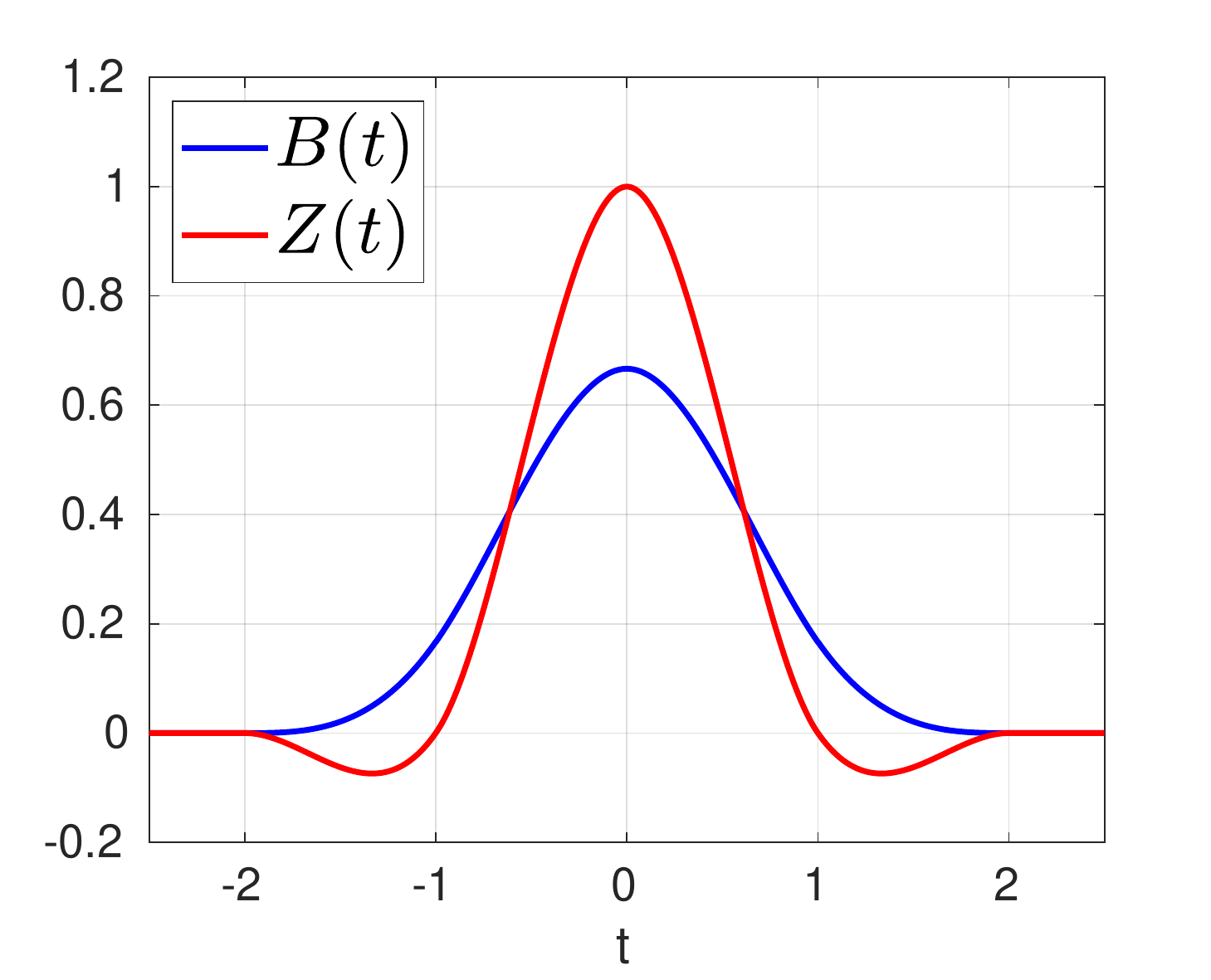}
        \caption{Standard Basis Function}
        \label{fig:basis_functions_plot}
    \end{subfigure}
	\begin{subfigure}[b]{0.48\columnwidth}
        \includegraphics[width=1.0\columnwidth]{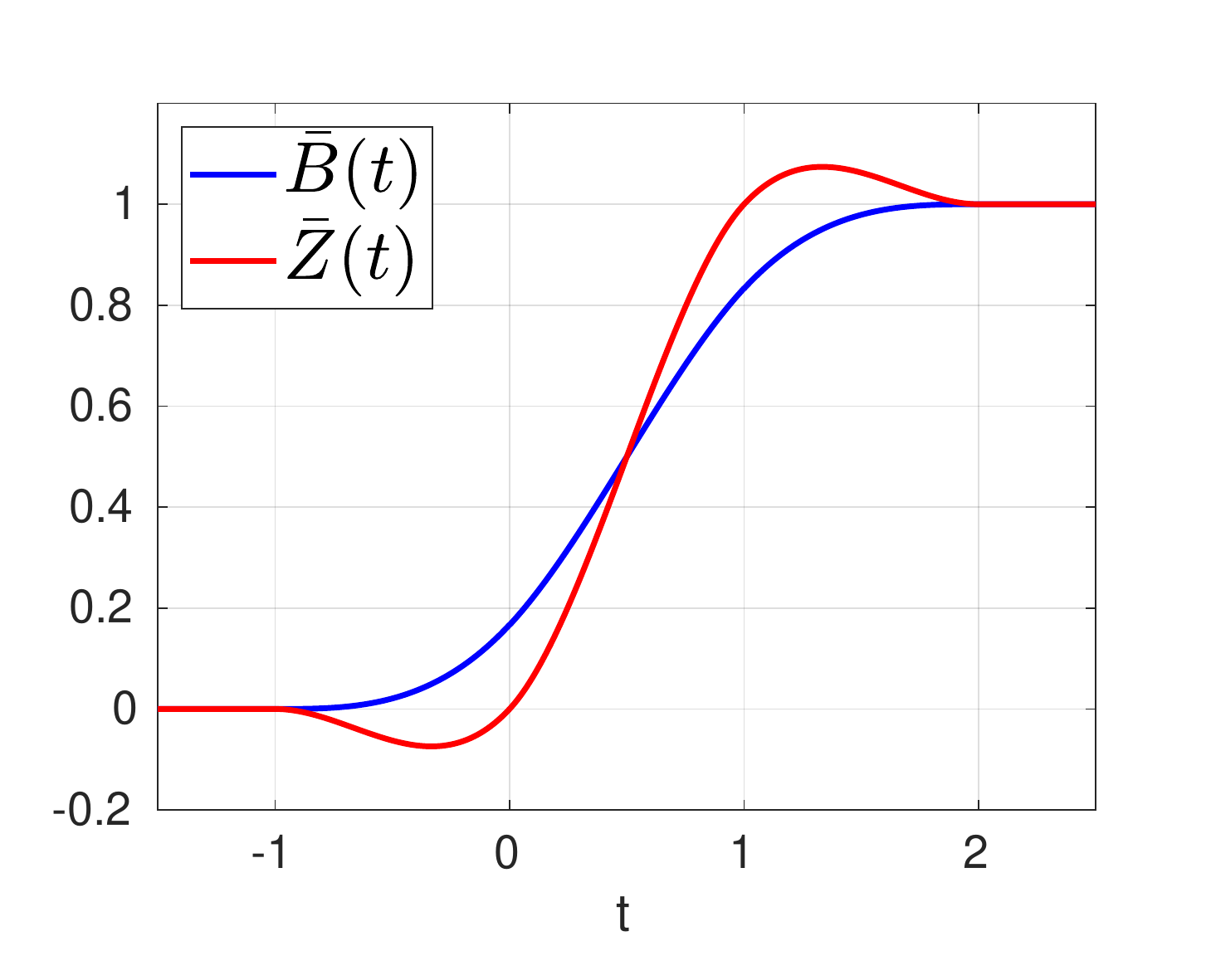}
        \caption{Cumulative basis Function}
        \label{fig:cum_basis_functions_plot}
    \end{subfigure}
    
    \begin{subfigure}[b]{0.99\columnwidth}
    	\includegraphics[width=1.0\columnwidth]{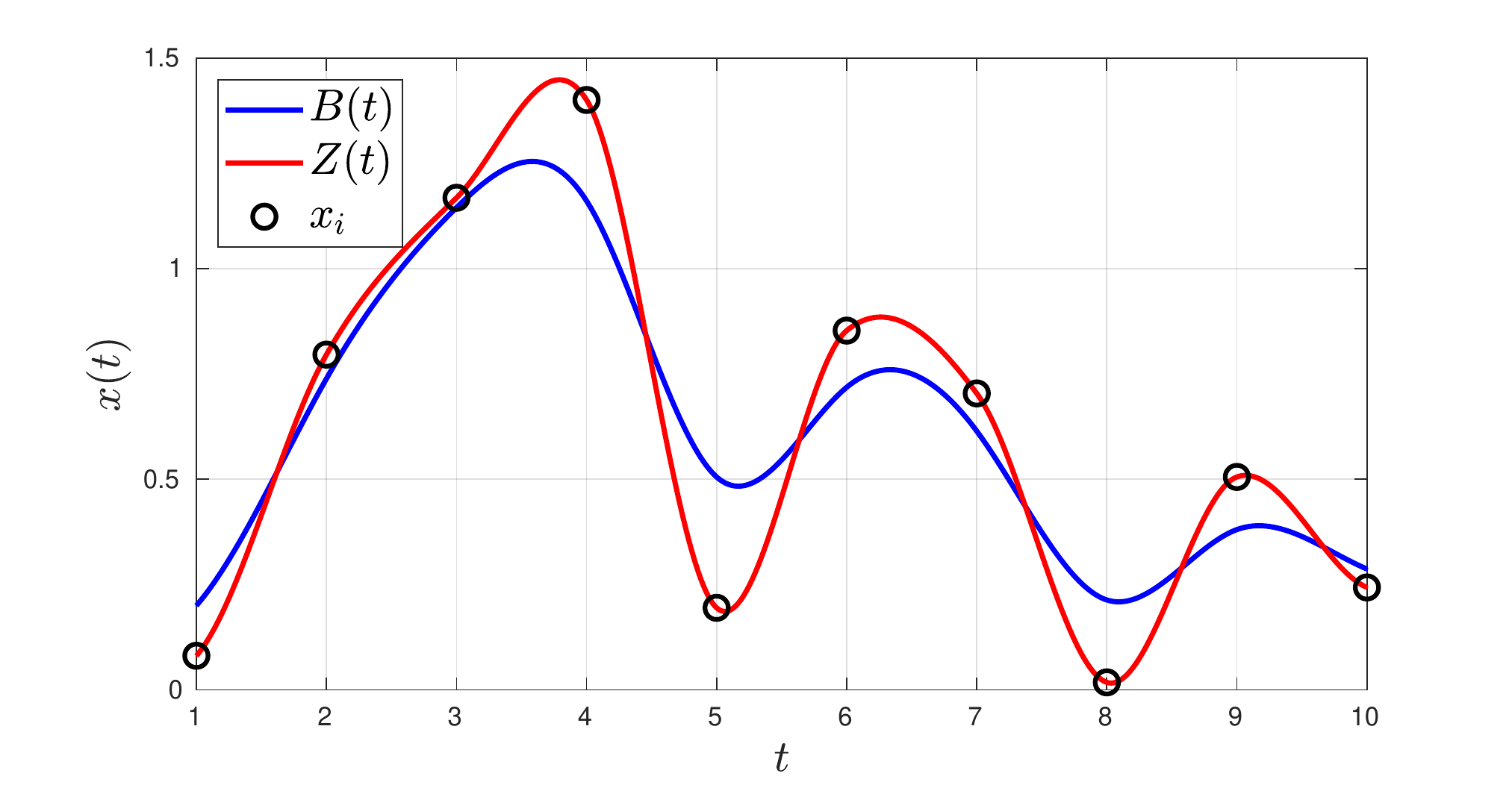}
    	\caption{Comparison between B-spline and Z-spline based interpolation}  
    	\label{fig:interpolation_comparison_plot}
    \end{subfigure}
    \caption{Comparison between the basis functions of B-splines and Z-splines. The plot in (a) shows the normal basis functions and (b) shows the corresponding cumulative basis function, while (c) shows a comparison of the interpolation result between B-splines and Z-splines using the same control points.}
\label{fig:basis_functions}
\end{figure}

Representing a trajectory as a continuous-time curve offers several advantages, for example the handling of sensors, which acquire data over a period of time, such as rolling-shutter cameras or LiDARs, as well as simplified fusion of data from multiple, possibly not time-synchronized sensors. 
One of the most popular representations for continuous-time trajectories are B-splines due to their simple parametric form and their local support properties.
In the literature, B-splines have been applied successfully to tackle visual-inertial \gls{slam} using rolling-shutter cameras \cite{Furgale:etal:IJRR2015}, \cite{Lovegrove:etal:BMVC2013}. 
As outlined in \cite{Lovegrove:etal:BMVC2013}, a standard B-spline curve of degree $k-1$ is defined by
\begin{equation}
\label{eq:standard_bspline}
\mathbf{x}(t) = \sum\limits_{i = 0}^{n}\mathbf{x}_{i} B_{i,k}(t) ~,
\end{equation}
where $\mathbf{x}_{i} \in \mathbb{R}^{N}$ are the control points of the spline at the times $t_{i}$ with $i \in [0, n]$ and $B_{i,k}(t)$ are the continuous time basis functions. 
By reorganizing Eq. \eqref{eq:standard_bspline}, the formulation can be brought into the cumulative form given by
\begin{equation}
\label{eq:cumulative_bspline}
\mathbf{x}(t) = \mathbf{x}_{0} \bar{B}_{0,k}(t)+ \sum\limits_{i = 1}^{n} (\mathbf{x}_{i} - \mathbf{x}_{i - 1}) \bar{B}_{i,k}(t) ~, 
\end{equation}
where $\bar{B}_{i,k}(t)$ represents the cumulative basis function and the relationship between the cumulative and the standard basis functions is described by
\begin{equation}
\label{eq:cumulative_basis_relation}
\bar{B}_{i,k}(t) = \sum\limits_{j= i}^{n}B_{j,k}(t) ~.
\end{equation}
The basis functions for B-splines can be easily computed using De Boor-Cox's recursive formula \cite{Cox:JAM1972}, \cite{DeBoor:JAT1972}. 
In this work in order to represent the continuous time trajectory, we propose to utilize a third order Z-spline representation introduced in \cite{Sagredo:Tercero:2003}. 
While Z-splines have the same local support properties as B-splines, their basis functions are defined as piece-wise polynomials. 
In particular, the third order basis function for the Z-spline are defined as
\begin{equation}
\label{eq:z_spline_basis_function}
Z(s) = \begin{cases}
1 - \frac{5}{2} s^{2} + \frac{3}{2} \lvert s \rvert^{3} & \lvert s \rvert \leq 1, \\
\frac{1}{2}(2 - \lvert s \rvert)^{2} (1 - \lvert s \rvert) & 1 < \lvert u \rvert  \leq 2, \\
0 & \lvert u \rvert > 2 ~,
\end{cases}
\end{equation}
where $s$ denotes a real-value scalar.
As in this work we employ cubic splines and utilize control points that are equally spaced in time (i.e. $t_{i+1} - t_{i} = \Delta t, \forall i \in [0, n - 1]$) for any $t \in [t_{i}, t_{i+1})$, exactly four control points are required, namely the ones at times $t_{i-1}, t_{i}, t_{i+1}$ and $t_{i+2}$.
Transforming the time $t$ into a local time $u(t) = \frac{t - t_{i}}{\Delta t}$, the interpolation using the cubic Z-spline can be written as:
\begin{equation}
\mathbf{x}(u) = Z(u + 1) \mathbf{x}_{i-1} + Z(u) \mathbf{x}_{i} + Z(u - 1) \mathbf{x}_{i+1} + Z(u - 2) \mathbf{x}_{i + 2} ~,
\end{equation}
with $u(t)$ denoted by $u$ for brevity.
Note that the definition of the cumulative form remains the same as in \eqref{eq:cumulative_bspline}. 
A comparison between the base functions and their cumulative forms is shown in \reffig{fig:basis_functions}. 
As evident in \reffig{fig:interpolation_comparison_plot}, in contrast to the B-splines, the Z-spline interpolation passes through the control points exactly.  
While the use of such a spline formulation on values in $\mathbb{R}^{N}$ is straightforward, the interpolation of rigid body transformations needs to be considered more carefully. 
The authors in \cite{Lovegrove:etal:BMVC2013} proposed to interpolate the 6 \gls{dof} pose using the cumulative formulation on $\mathbb{SE}3$, however, here we perform the interpolation of the rotation on $\mathbb{SO}3$ and the translation parts on $\mathbb{R}^{3}$ separately. 
This was already advocated in \cite{Kim:etal:SIGGRAPH1995}, as such a split representation removes the coupling of the translation and the rotation and avoids artifacts on the translation-interpolation during phases with large rotational velocities. 
Therefore, the interpolated translation is given by
\begin{equation}
\label{eq:translation_interpolation}
\mathbf{p}(u) = \mathbf{p}_{i-1} \bar{Z}(u+1) + \sum\limits_{j=0}^{2} (\mathbf{p}_{i+j} - \mathbf{p}_{i + j - 1}) \bar{Z}(u - j) ~.
\end{equation}
Following the approach of \cite{Kim:etal:SIGGRAPH1995}, the interpolated rotation is computed by
\begin{equation}
\label{eq:rotation_interpolation}
\mathbf{q}(u) = \mathbf{q}_{i - 1}^{\bar{Z}(u + 1)} \prod\limits_{j=0}^{2} \mathrm{exp}\left( \mathrm{log}(\mathbf{q}_{i + j - 1}^{-1} \mathbf{q}_{i + j}) \bar{Z}(u - j) \right) ~,
\end{equation}
where $\mathbf{q}^{\lambda} = \mathrm{exp}(\lambda \, \mathrm{log}(\mathbf{q}))$. 
Essentially, the operations in \eqref{eq:rotation_interpolation} correspond to a mixture of SLeRP interpolations \cite{Shoemake:Ken:1985} within the local support window of the spline.

\section{Method}\label{sec:method}
\subsection{System Overview}
\label{ssec:system_overview}
\begin{figure*} [ht]
\vspace{0mm}
		\centering
		\includegraphics[angle=0,width=0.95\textwidth]{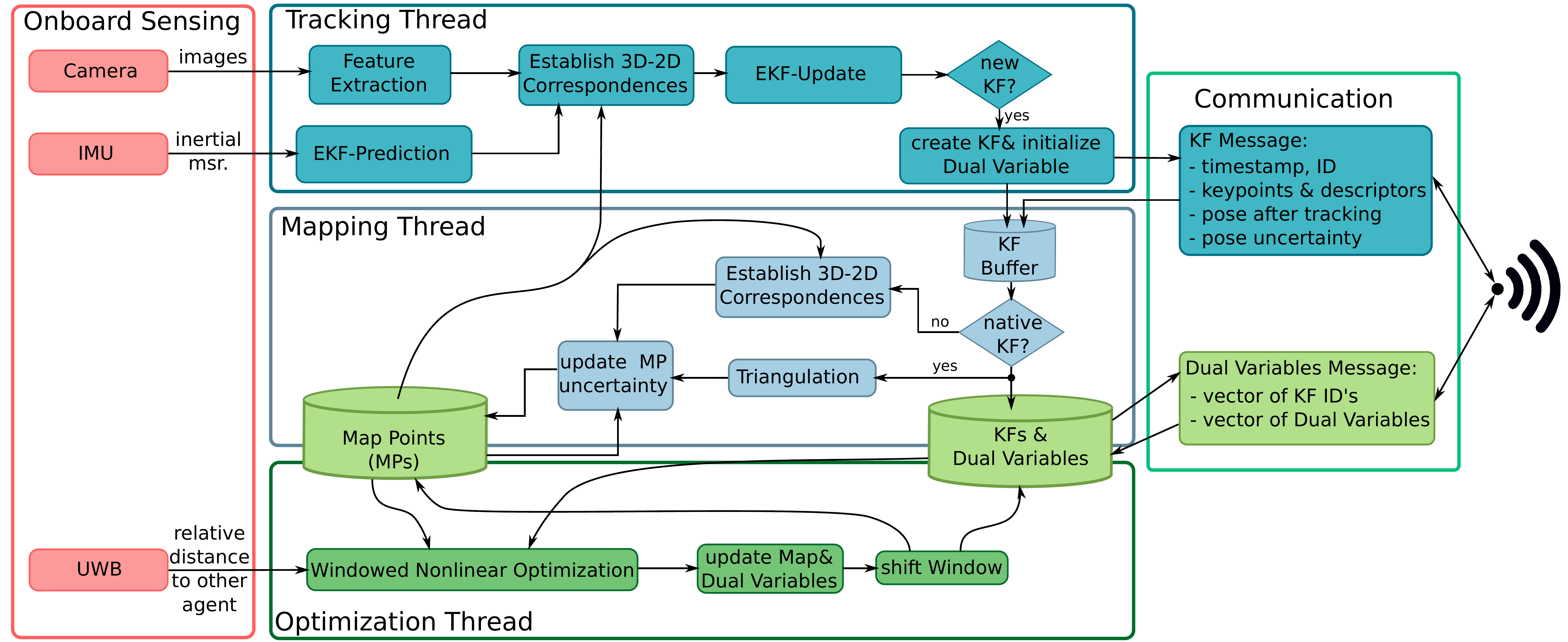}
		\caption{Schematic of the processes running on each of the two agent for the proposed system  mainly comprising of frame-wise tracking, mapping and non-linear optimization. Consistency between the two agents is achieved by communicating Keyframe (KF) data and Dual Variables associated to the KF poses via wireless communication.}
\vspace{0mm}
\label{fig:system_schematics}
\end{figure*}

The system performing distributed, collaborative \gls{slam} is designed to be run as two identical instances on two machines, called `agents', communicating with each other over a wireless connection. 
An overview of the essential parts of one such instance is shown in \reffig{fig:system_schematics}. 
The functionality of each instance is partitioned into three main threads: tracking, mapping, and optimization. \\ 
In the \textit{tracking thread}, the image observations are fused together with \gls{imu} readings in order to localize against the current state of the 3D \glspl{mp}. 
The fusion is performed by the means of an \gls{ekf}, which allows for a computationally efficient and low-latency frame-wise pose tracking. 
The tracked frame is further processed to decide whether a \gls{kf} should be created or not. 
In the proposed system this decision is purely based on time constraints, i.e. if the time difference between the last \gls{kf} and the current frame is larger than a threshold, the current frame is marked to become a \gls{kf}. 
This scheme is motivated by the simplicity in the book-keeping, but also by the fact that we use the \gls{kf} poses as base poses in the spline representation of the trajectory in the back-end optimization, hence, using uniform sampling between the \glspl{kf} reduces the complexity of the interpolation. 
When a \gls{kf} is newly created at one agent, the relevant information such as the 2D keypoints, descriptor data as well pose related information gets immediately communicated to the other agent. 
In the following the \glspl{kf} created on its own by an agent are referred to as native \glspl{kf}, while \glspl{kf} received via communication but created by the other agent are referred to as non-native \glspl{kf}.
Both, native- as well as non-native \glspl{kf} are passed to the mapping part of the pipeline.
%
In the \textit{mapping thread}, we try to establish new 3D \glspl{mp} via triangulation between the newest native \gls{kf} and a selected subset of established \glspl{kf}.
In order to reduce the required book-keeping for the map maintenance, the system is designed such that each agent can hold its own set of map points, avoiding complex synchronization between the two agents.
Besides the creation of new \glspl{mp}, the mapping thread is also responsible to update the uncertainty estimates of the \glspl{mp} and establish correspondences between the existing \glspl{mp} and non-native \glspl{kf}.
In order to avoid extensive locking of data during these operations, the interface between the tracking- and the mapping-thread copies the last state of the \glspl{mp} to be used for tracking.
While the \textit{mapping thread} initializes the \gls{mp} positions and maintains an estimate of their uncertainty, we utilize nonlinear optimization to update the \gls{mp} positions and fuse \gls{uwb} distance measurements between the two agents into the estimation. \\
As we have two instances of the map estimated, one on each agent, in the \textit{optimization thread} we make use of the asynchronous \gls{admm} introduced in \refsec{ssec:async_admm} to ensure that both agents converge to a common trajectory estimate by the means of a pose consensus, enforced by exchanging a set of dual variables.
In order to keep the computational complexity of the optimization bounded, we only keep a fixed sized window of \glspl{kf} in the Map, whose size is maintained after every round of optimization.

\subsection{EKF-based Pose Tracking}
\label{ssec:ekf_pose_tracking}
In order to enable both robust and timely tracking of each robot's pose, we designed an \gls{ekf} fusing the \gls{imu} information together with 3D-2D measurements against the current state of the \glspl{mp}. 
As the \gls{mp} positions are conceptually considered as given in the tracking thread, the \gls{ekf} state $\mathcal{X}_{tr}$ is chosen as follows:
\begin{equation}
\label{eq:ekf_state}
\mathcal{X}_{tr} \defeq \begin{bmatrix}
\mathbf{q}_{WM} & \mathbf{q}_{MS} & \mathbf{p}_{MS} & \prescript{}{S}{\mathbf{v}} & \mathbf{b}_{a} & \mathbf{b}_{\omega}
\end{bmatrix} ~,
\end{equation}
where $\mathbf{q}_{WM}$ denotes the rotation (in quaternion form) of this agent's map origin into the gravity aligned world frame, $[\mathbf{q}_{MS}, \mathbf{p}_{MS}]$ corresponds to the pose of the \gls{imu} frame in the map, $_{S}\mathbf{v}$ denotes the robocentric velocity of the \gls{imu} frame, and $\mathbf{b}_{a}$ and $\mathbf{b}_{\omega}$ denote the accelerometer- and the gyroscope biases, respectively.
The covariance associated to the state $\mathcal{X}_{tr}$ is denoted as $\pmb{\Sigma}_{tr}$. While rotations are parameterized as quaternions, the gravity rotation ($\mathbf{q}_{WM}$)  only has two \gls{dof} (roll and pitch), whereas the pose ($\mathbf{q}_{MS}$) has $3$DoF, resulting in $\pmb{\Sigma}_{tr} \in \mathbb{R}^{17 \times 17}$. 
As commonly employed in filtering based visual-intertial frameworks, such as in \cite{Bloesch:etal:IJRR2017}, we use the \gls{imu} readings to propagate the filter state. 
The  acceleration $\prescript{}{S}{\tilde{\mathbf{a}}}_{S}$ and gyroscope measurements $\prescript{}{S}{\tilde{\pmb{\omega}}}_{WS}$ are modeled to be the true acceleration $\prescript{}{S}{\mathbf{a}}_{S}$ and rotational velocity $\prescript{}{S}{\pmb{\omega}}_{WS}$ affected by both noise and biases:
\begin{align}
\label{eq:imu_model}
\prescript{}{S}{\mathbf{a}}_{S} &= \prescript{}{S}{\tilde{\mathbf{a}}}_{S} - \mathbf{b}_{a}  - \mathbf{w}_{a} \\
\prescript{}{S}{\pmb{\omega}}_{WS} &= \prescript{}{S}{\tilde{\pmb{\omega}}}_{WS} - \pmb{b}_{\omega} - \mathbf{w}_{\omega} ~,
\end{align}
where $\mathbf{w}_{a}, \mathbf{w}_{\omega}$ are zero-mean Gaussian noise variables acting on the acceleration and the rotational velocity measurements, respectively. 
Using the \gls{imu} readings, the continuous time behavior of the system can be described by
\begin{align}
\label{eq:continuous_ekf_model}
\dot{\mathbf{q}}_{WM} &= \mathbf{w}_{g} \\
\dot{\mathbf{q}}_{MS} &= -\mathbf{q}_{MS}^{-1}(\mathbf{w}_{g}) + _{S}\pmb{\omega}_{WS}  \\
\dot{\mathbf{p}}_{MS} &=   - \prescript{}{S}{\pmb{\omega}}_{WS}^{\times} \mathbf{p}_{MS} + \prescript{}{S}{\mathbf{v}}_{S}\\
\prescript{}{S}{\dot{\mathbf{v}}}_{S} &=  -\prescript{}{S}{\pmb{\omega}}_{WS}^{\times} \prescript{}{S}{\mathbf{v}}_{S} + \prescript{}{S}{\mathbf{a}}_{S} + \mathbf{q}_{MS}^{-1} \circ \mathbf{q}_{WM}^{-1} (\prescript{}{W}{\mathbf{g}})  \\
\dot{\mathbf{b}}_{a} &=  \mathbf{w}_{b_{a}} \\
\dot{\mathbf{b}}_{\omega} &=  \mathbf{w}_{b_{\omega}} ~, 
\end{align}

where $\mathbf{w}_{b_{a}}$ and $\mathbf{w}_{b_{g}}$ are white Gaussian noise processes modeling the time variation of the accelerometer and gyroscope biases and $\mathbf{\omega}^{\times}$ is denoting the skew symmetric matrix constructed from the tuple $\pmb{\omega}$.
The term $\prescript{}{W}{\mathbf{g}}$ denotes the gravity vector in the inertial frame and the Gaussian noise process $\mathbf{w}_{g}$ models a time variation of the rotation of the map with respect to the inertial frame $W$. 
The time continuous equations in our implementation are transformed to a set of discrete prediction equations using an Euler forward integration scheme as proposed in \cite{Bloesch:etal:IJRR2017} resulting in:
\begin{align}
\hat{\mathbf{q}}_{WM}^{k+1} &= \mathbf{q}_{WM}^{t} \label{eq:discrete_ekf_prediction1}\\
\hat{\mathbf{q}}_{MS}^{k+1} &= \mathbf{q}_{MS}^{k}\boxplus (\Delta t \prescript{}{S}{\pmb{\omega}}_{WS}^{k}) \label{eq:discrecte_ekf_boxplus}  \\
\hat{\mathbf{p}}_{MS}^{k+1} &= \mathbf{p}_{MS}^{k} + \Delta t \mathbf{q}_{MS}^{k} (\prescript{}{S}{\mathbf{v}}_{S}^{k}) \\
\prescript{}{S}{\hat{\mathbf{v}}}_{S}^{k+1} &= \prescript{}{S}{\mathbf{v}}_{S}^{k} + \Delta t \left( (\mathbf{q}_{WM}^{k} \circ \mathbf{q}_{MS}^{k})^{-1}(\prescript{}{W}{\mathbf{g}}) \right. +  \nonumber \\
  & \left. \prescript{}{S}{\mathbf{a}}_{S}  -(\prescript{}{S}{\pmb{\omega}}_{WS}^{k})^{\times} \prescript{}{S}{\mathbf{v}}_{S}^{k}\right) \\
\hat{\mathbf{b}}_{a}^{k+1} &= \mathbf{b}_{a}^{k} \\
\hat{\mathbf{b}}_{g}^{k+1} &= \mathbf{b}_{g}^{k}  ~. \label{eq:discrete_ekf_prediction2}
\end{align}
Note that in Eq. \eqref{eq:discrecte_ekf_boxplus} we use the boxplus operator ($\boxplus$), which generalizes the functionality of addition for quantities which are not in a vector space \cite{Hertzberg:etal:INFUS2011}.
In this case the $\boxplus$ operator is used on rotations and is briefly outlined in Appendix \ref{ssec:rotation_calculus}.
Based on the discrete prediction equations \eqref{eq:discrete_ekf_prediction1}-\eqref{eq:discrete_ekf_prediction2}, we propagate the state covariance as follows:
\begin{equation}
\label{eq:ekf_covariance_prediction}
\hat{\pmb{\Sigma}}_{tr}^{k+1} = \mathbf{F} \pmb{\Sigma}_{tr}^{k} \mathbf{F}^{T} + \mathbf{G} \mathbf{W} \mathbf{G}^{T} ~,
\end{equation}
where $\mathbf{F}$ is the jacobian of the prediction step with respect to the state, $\mathbf{G}$ is the jacobian with respect to the process noise $\mathbf{w}_{*}$ and $\mathbf{W}$ is the covariance of the process noise in matrix form. 
The analytical expressions and the form of $\mathbf{G}, \mathbf{F}$ and $\mathbf{W}$ are provided in Appendix \ref{ssec:jacobians_pose_tracking_ekf}.\\
For updating the \gls{ekf}-state, we use projective correspondences to the current position estimates of the \glspl{mp} as measurements. 
In order to establish these correspondences, we use the predicted pose $\hat{\mathbf{q}}_{MS}^{k+1}, \hat{\mathbf{p}}_{MS}^{k+1}$ and project all \glspl{mp} $\prescript{}{M}{\mathbf{m}}_{i}$ into the camera frame:
\begin{equation}
\label{eq:landmark_projection}
{\mathbf{z}_{proj}}_{i} = \pi(\mathbf{T}_{CS} \hat{\mathbf{T}}_{MS}^{-1} \prescript{}{M}{\mathbf{m}}_{i}) ~,
\end{equation}
where the function $\pi(\cdot)$ denotes the projection function (including distortion).
The transformation $\mathbf{T}_{CS}$ between the \gls{imu} and the camera  is considered to be known and can be obtained from calibration.
In order to associate the 2D keypoints to the projected \glspl{mp}, the descriptors in a small radius around the projection ${\mathbf{z}_{proj}}_{i}$ are matched against the \gls{mp}'s descriptors.
After a first data association step, we perform an outlier rejection step using 3D-2D RANSAC \cite{Kneip:etal:CVPR2011}. 
Using the remaining inlier correspondences, the reprojection residuals given by
\begin{equation}
\label{eq:ekf_reprojection_residual}
\mathbf{y}_{i,j} = \tilde{\mathbf{z}}_{j} - \mathbf{z}_{proj_{i}}
\end{equation}
are constructed, where $\tilde{\mathbf{z}}_{j}$ is the pixel coordinates of the 2D keypoint $j$ in the image space that was associated with \gls{mp} $i$. 
By stacking all reprojection residuals to form the residual vector $\mathbf{y}$, we can formulate the innovation covariance as
\begin{equation}
\label{eq:ekf_innovation_covariance}
\mathbf{S} = \mathbf{H} \hat{\pmb{\Sigma}}_{tr} \mathbf{H}^{T} + \mathbf{R} ~,
\end{equation}
where $\mathbf{H}$ is the jacobian matrix of the residual with respect to the \gls{ekf} state and $\mathbf{R}$ is the measurement covariance obtained by stacking the individual measurement covariance $\mathbf{R}_{i,j}$ associated to $\mathbf{y}_{i,j}$.
As the \glspl{mp} are not part of the \gls{ekf} state, we use the \gls{mp} uncertainty $\pmb{\Sigma}_{p_{i}}$, estimated as described in \refsec{ssec:mapping}, in order to inflate the measurement uncertainty by projecting it onto the image plane:
\begin{equation}
\label{eq:ekf_residual_covariance}
\mathbf{R}_{i,j} = \mathbf{H}_{\mathbf{T}_{MS},i,j} \pmb{\Sigma}_{p_{i}}  \mathbf{H}_{\mathbf{T}_{MS},i,j}^{T} + 
\begin{bmatrix}
\sigma_{obs}^{2} & 0 \\
0 & \sigma_{obs}^{2}
\end{bmatrix} ~,
\end{equation}
where $\mathbf{H}_{\mathbf{T}_{MS},i,j}$ is the part of the residual jacobian $\mathbf{H}$ corresponding to the pose $\mathbf{T}_{MS}$ and $\sigma_{obs}$ is the keypoint uncertainty, which in our case is only dependent on the octave that this keypoint was detected.
%
Leveraging the computed innovation covariance, we employ a Mahalanobis distance based outlier rejection in order to exclude additional outliers that slipped through the first RANSAC step.
Using the remaining correspondences, the Kalman gain can be computed by
\begin{equation}
\label{eq:ekf_kalman_gain}
\mathbf{K} = \mathbf{H} \hat{\pmb{\Sigma}}_{tr} \mathbf{H}^{T} \mathbf{S}^{-1} ~.
\end{equation}
The computed gain is utilized to update the state variables and the associated covariance as follows:
\begin{align}
\label{eq:ekf_update}
\mathcal{X}_{tr} &= \hat{\mathcal{X}}_{tr} \boxplus (-\mathbf{K} \mathbf{y})  \\
\pmb{\Sigma}_{tr} &= (\mathbf{I}_{17 \times 17} - \mathbf{K}\mathbf{H}) \hat{\pmb{\Sigma}}_{tr} ~,
\end{align}
where the notation in Eq. \eqref{eq:ekf_update} indicates that the $\boxplus$ operator is applied for the appropriate states.

\subsection{Mapping}
\label{ssec:mapping}
The \textit{mapping thread} in the proposed pipeline is responsible to generate new \glspl{mp} and to maintain an estimate of the uncertainty in their positions. 
\subsubsection{Initialization of new Map Points}
\label{sssec:initialization_map_points}
The initialization of new \glspl{mp} is performed by triangulating of 2D correspondences between the most recent native \gls{kf} (target) and the \glspl{kf} that are already in the Map (candidates).
As an exhaustive correspondence search against all established \glspl{kf} would be computationally too expensive, instead, we propose to limit the search to a small subset of candidate \glspl{kf}.
In order to select such a subset, we assign a penalty value $s_{j}$ to every established \gls{kf} $j$ based on the relative viewpoint with respect to the target \gls{kf} $i$ as follows:
\begin{equation}
\label{eq:keyframe_score}
s_{j} := \beta_{v,j} \cdot w(\alpha_{v,j}, a_{v}, b_{v}, c_{v}) + \beta_{t,j} \cdot w(\alpha_{t,j}, a_{t}, b_{t}, c_{t})~,
\end{equation}
where $\beta_{v,j}, \beta_{t,j}$ are weights chosen such that non-native \glspl{kf} are preferred in order to establish a stronger constraints across the agents, i.e. the weights for native \glspl{kf} are chosen 10 times larger.
The value $\alpha_{v,j}$ is the angle between the camera axes of \gls{kf} j and \gls{kf} i, and $\alpha_{t,j}$ is defined as the triangulation angle between the \glspl{kf} for a given scene depth.
The parameters $a_{i}, b_{i}, c_{i} \text{ with } i \in \{v, t \}$ are internal parameters of the weighting function $w$.
In order to approximate the unknown scene depth, we compute the median distance of the \glspl{mp} seen in the most recent couple of \glspl{kf}.
The weighting function $w(x, a, b, c)$ is chosen in the form of a tolerant loss function:
\begin{align}
\label{eq:keyframe_score_weighting}
w(x,a,b,c) &= b \cdot \mathrm{log}\left(1 + \mathrm{exp} \left(\frac{(x - c)^{2} - a}{b}  \right)\right) - c_{0} \\
c_{0} &=  b \cdot \mathrm{log}\left(1.0 + \mathrm{exp} \left(\frac{-a}{b} \right) \right) ~. \nonumber
\end{align}
%
Using the candidate \glspl{kf} with the smallest score, we sequentially perform a brute force descriptor matching followed by a 2D-2D RANSAC-based outlier rejection.
For the obtained inlier correspondences we perform a SVD-based linear triangulation to obtain the 3D position of the new map points $_{M}{\mathbf{m}}_{i,j}$  using the \glspl{kf} pose estimates $\mathbf{T}_{MS_{i}}, \mathbf{T}_{MS_{j}}$.
Leveraging the estimated covariances of the \glspl{kf} poses $\pmb{\Sigma}_{\mathbf{T}_{MS_{i}}}, \pmb{\Sigma}_{\mathbf{T}_{MS_{j}}}$, we estimate the \gls{mp} uncertainty $\pmb{\Sigma}_{\prescript{}{M}{\mathbf{m}}_{i,j}}$ as:
\begin{equation}
\label{eq:initial_map_point_uncertainty}
\pmb{\Sigma}_{\prescript{}{M}{\mathbf{m}}_{i,j}} = \left( \mathbf{J}_{i,j} \mathbf{W}_{i,j} \mathbf{J}_{i,j}  \right)^{-1} \left[1,3 ; 1,3 \right]~,
\end{equation}
where $\mathbf{J}_{i,j}$ is the jacobian of the reprojected \glspl{mp} and the poses.
The form and the analytical expression of $\mathbf{J}_{i,j}$ is provided in Appendix \ref{ssec:jacobians_pose_tracking_ekf}.
%
The matrix $\mathbf{W}_{i,j}$ is the corresponding information matrix and is given by
\begin{equation}
\label{eq:initial_map_points_information}
\mathbf{W}_{i,j} = \begin{bmatrix}
\frac{1}{\sigma_{obs}^{2}} \mathbf{I}_{2\times 2} & \mathbf{0} & \mathbf{0} \\
\mathbf{0} & \pmb{\Sigma}_{\mathbf{T}_{MS_{i}}}^{-1} & \mathbf{0} \\
\mathbf{0} & \mathbf{0} & \pmb{\Sigma}_{\mathbf{T}_{MS_{j}}}^{-1}
\end{bmatrix} ~.
\end{equation}

\subsubsection{Estimation of the Map Points' uncertainty}
\label{sssec:estimation_map_point_uncertainty}
As introduced in \refsec{ssec:ekf_pose_tracking}, we utilize the uncertainty of \glspl{mp} in the update of the pose tracking.
In order to obtain an approximation of the uncertainty for each \gls{mp} with limited computational effort, we employ an independent \gls{ekf} per \gls{mp}.
For the initialization of the \gls{ekf}s' state, we use the triangulated positions along with the covariance as computed by Eq. \eqref{eq:initial_map_point_uncertainty}. 
As the \gls{ekf}'s state represents the \gls{mp} position in $M$, the prediction step from timestep $k$ to $k+1$ is trivial, as the \glspl{mp} are assumed to be static:
\begin{equation}
\label{eq:map_point_ekf_prediction_state}
\prescript{}{M}{\hat{\mathbf{m}}}^{k+1}_{i} = \prescript{}{M}{\mathbf{m}}_{i}^{k} ~,
\end{equation}
while for the covariance we add diagonal noise to account for missing or removed observations due to the sliding window approach:
\begin{equation}
\label{eq:map_point_ekf_prediction_cov}
\hat{\pmb{\Sigma}}_{m_{i}}^{k+1} = \pmb{\Sigma}_{m_{i}}^{k} + \sigma_{m}^{2}  \cdot \mathbf{I}_{3\times 3} ~,
\end{equation}
where $\sigma_{m}$ is the noise parameter modeling the \gls{mp}'s changes in position (e.g. during optimization) and in our implementation was chosen to be $0.2m$.
In the update step, we use the re-projection error $\mathbf{y}_{i}$ of the established 2D-3D correspondences, independent of whether the \gls{kf} used is a native or not.
In order to include the uncertainty of the \gls{kf} pose into the estimation, a similar operation as in Eq. \eqref{eq:ekf_residual_covariance} is employed:
\begin{equation}
\label{eq:map_point_ekf_meas_covariance}
\mathbf{R}_{\mathbf{p}_{y,i}} = \mathbf{H}_{\mathbf{T}_{MS}} \hat{\pmb{\Sigma}}_{\mathbf{T}_{MS}} \mathbf{H}_{\mathbf{T}_{MS}}^{T} + \sigma_{obs}^{2} \cdot \mathbf{I}_{2 \times 2} ~,
\end{equation}
where the Jacobian $\mathbf{H}_{\mathbf{T}_{MS}}$ is given by
\begin{equation}
\label{eq:map_point_ekf_meas_jac}
\mathbf{H}_{\mathbf{T}_{MS}} = \frac{\partial \mathbf{y}_{i}}{\partial \mathbf{T}_{MS}} ~.
\end{equation}
With the resulting measurement noise from Eq. \eqref{eq:map_point_ekf_meas_covariance}, the innovation covariance $\mathbf{S}_{\mathbf{y}_{i}}$ can be computed as
\begin{equation}
\label{eq:map_point_ekf_innoviation}
\mathbf{S}_{\mathbf{y}_{i}} = \mathbf{H}_{\mathbf{p}_{i}} \hat{\pmb{\Sigma}}_{\mathbf{p}_{i}} \mathbf{H}_{\mathbf{p}_{i}}^{T} + \mathbf{R}_{\mathbf{p}_{y,i}} ~.
\end{equation} 
Using the resulting Kalman Gain $\mathbf{K}_{\mathbf{y}_{i}} = \pmb{\Sigma}_{\mathbf{p}_{i}}\mathbf{H}_{\mathbf{p}_{i}}  \mathbf{S}_{\mathbf{y}_{i}}^{-1}$, the updated \gls{mp} position is given by
\begin{equation}
\label{eq:map_point_ekf_state_update}
\mathbf{p}_{i} = \hat{\mathbf{p}}_{i} - \mathbf{K}_{\mathbf{y}_{i}} \mathbf{y}_{i} ~,
\end{equation}
while the corresponding \gls{mp}'s covariance is given by 
\begin{equation}
\label{eq:map_point_ekf_covariance_update}
\pmb{\Sigma}_{\mathbf{p}_{i}} = \hat{\pmb{\Sigma}}_{\mathbf{p}_{i}} - \mathbf{K}_{\mathbf{y}_{i}} \mathbf{S}_{\mathbf{y}_{i}} \mathbf{K}_{\mathbf{y}_{i}}^{T} ~.
\end{equation}
Note that the described \gls{ekf} is mainly employed to estimate the uncertainty of the \glspl{mp}, therefore, the state update in Eq. \eqref{eq:map_point_ekf_state_update} is only performed until the \gls{mp} has seen the first update from the Nonlinear Optimization as described in \refsec{ssec:optimization}.

\subsection{Distributed Optimization Back-End}
\label{ssec:optimization}
The optimization back-end constitutes the core element of the system and is responsible for the fusion of the \gls{uwb} distance-measurements together with the visual measurements and maintaining a consistent estimate among the two agents.
In this section, we first introduce the optimization objective assuming a centralized system followed by the undertaken steps to optimize the objective in a decentralized fashion.
\subsubsection{Centralized Objective Function}
\label{sssec:centralized_objective}
The optimization variables in the back-end consist of the \gls{kf} poses inside a fixed window size of $N$ and the corresponding $M$ \glspl{mp} visible in these \glspl{kf}:
\begin{equation}
\label{eq:optimization_state}
\resizebox{.999 \columnwidth}{!} 
{$
\mathcal{X} := [ \underbrace{\mathbf{T}_{MS_{A}}^{1}, \cdots \mathbf{T}_{MS_{A}}^{N}}_{\text{agent A's \glspl{kf}}}, \underbrace{\mathbf{T}_{MS_{B}}^{1}, \cdots, \mathbf{T}_{MS_{B}}^{N}}_{\text{agent B's \glspl{kf}}}, \underbrace{{}_{M}\mathbf{m}^{1}, \cdots, {}_{M}\mathbf{m}^{M}}_{\text{Map Points}} ] ~.
$}
\end{equation}
Over these variables, we can define the optimization objective as:
\begin{equation}
\label{eq:optimization_objective}
\resizebox{.999 \columnwidth}{!} 
{$
f(\mathcal{X}) := \sum\limits_{i \in \mathcal{K}}\sum\limits_{j \in \mathcal{M}(i)} \delta_{c}\left(\mathbf{y}_{proj_{i,j}}^{T} \mathbf{W}_{r}^{i,j} \mathbf{y}_{proj_{i,j}}^{T} \right) + \sum\limits_{u \in \mathcal{D}} \delta_{c} \left( \frac{1}{\sigma_{d}^{2}}{e_{d}^{u}}^{2} \right) ~,
$}
\end{equation}
where the set $\mathcal{K}$ indicates all \glspl{kf} $i$, native and non-native, that are currently inside the sliding window and accordingly, $\mathcal{M}(i)$ indicates all the \glspl{mp} that are visible in \gls{kf} $i$.
The function $\delta_{c}(\cdot)$ denotes a robust loss function, in our case the Cauchy loss function, introduced to reduce the influence of outliers.
The terms $\mathbf{y}_{proj_{i,j}}$ are the reprojection residuals as defined in Eq. \eqref{eq:landmark_projection} and \eqref{eq:ekf_reprojection_residual} and the corresponding weights are $\mathbf{W}_{r}^{k,j} = 1/\sigma_{obs}^{2} \cdot \mathbf{I}_{2\times 2}$.
The set of relative distance measurements with standard deviation $\sigma_{d}$ between the two agents is denoted by $\mathcal{D}$, while the corresponding residual terms are given by
\begin{equation}
\label{eq:relative_distance_residual}
\resizebox{.999 \columnwidth}{!} 
{$
e_{d}^{u} := \rVert \mathbf{q}_{MS_{A}}(t) \mathbf{p}_{U_{A}} + \mathbf{p}_{MS_{A}}(t) - \mathbf{q}_{MS_{B}}(t) \mathbf{p}_{U_{B}} - \mathbf{p}_{MS_{B}}(t) \rVert - d_{meas}^{u} ~,
$}
\end{equation}
where $d_{meas}^{u}$ corresponds to the distance measurement taken at time $t$, and $\mathbf{p}_{U_i}, i \in \{ A, B \}$ is the \gls{uwb} antenna offset expressed in the corresponding \gls{imu} frame.
The interpolated poses $\mathbf{p}_{MS_{i}}(t), \mathbf{q}_{MS_{i}}(t), i \in \{ A,B \}$ are computed following \eqref{eq:translation_interpolation} and Eq. \eqref{eq:rotation_interpolation}, while the base poses correspond to the \gls{kf} poses surrounding the timestamp $t$.

\subsubsection{Distributed Optimization}
\label{sssec:distributed_optimization}
To avoid a complex synchronization effort between the two agents in order to obtain a unique, common map, we allow each agent to hold its own version of the map and, instead, propose to use an \gls{admm} based distributed optimization scheme to obtain a common trajectory estimate across both agents.
In order to optimize $f(\mathcal{X})$ in a distributed fashion, the problem in Eq. \eqref{eq:optimization_objective} needs to be brought into the form of Eq. \eqref{eq:arock_admm_base}.
Based on the need for both agents' trajectories in \eqref{eq:relative_distance_residual} and the absence of a shared map across the agents, we split the problem by their trajectories, resulting in the following distributed state:
%
\begin{equation}
\label{eq:optimization_state_distributed}
\resizebox{.999 \columnwidth}{!} 
{$
\mathcal{X}_{i} := [ \underbrace{\mathbf{T}_{MS_{i}}^{1}, \cdots \mathbf{T}_{MS_{i}}^{N}}_{\text{native KFs}}, \underbrace{\mathbf{T}_{MS_{\ulcorner i}}^{1}, \cdots, \mathbf{T}_{MS_{\ulcorner i}}^{N-l}}_{\text{non-native KFs}}, \underbrace{{}_{M}\mathbf{m}^{1}, \cdots, {}_{M}\mathbf{m}^{M_{i}}}_{\text{Map Points}} ], i \in \{ A, B \} ~,
$}
\end{equation}
where the notation $\ulcorner i$ indicates the opposite index, i.e. $i = A \Rightarrow \ulcorner i = B$.
The parameter $l \geq 0$ is used to represent a lag between the creation of a \gls{kf} on one agent until it is available to the other one, e.g. caused by network delays.
To ensure consistency between the trajectories on both agents, the constraint in Eq. \eqref{eq:arock_admm_base} for the state as in Eq. \eqref{eq:optimization_state_distributed} is given by
\begin{equation}
\label{eq:pose_consensus_constraint}
\mathbf{T}_{MS_{A,A}}^{i} = \mathbf{T}_{MS_{A,B}}^{i}, \quad \mathbf{T}_{MS_{B,A}}^{i} = \mathbf{T}_{MS_{B,B}}^{i} , \quad \forall i \in [1, N - l] ~. 
\end{equation}
Using these constraints, the centralized problem in \eqref{eq:optimization_objective} can be written in the form of \eqref{eq:arock_min}
\begin{equation}
\label{eq:distributed_objective}
\mathcal{X}_{i}^{k+1} = \underset{\mathcal{X}_{i}}{\mathrm{arg\, min}} f(\mathcal{X}_{i}) + \sum\limits_{j = 1}^{N}\lVert \mathbf{e}_{c}^{i,j} \rVert^{2} + \sum\limits_{j = 1}^{N-l} \lVert \mathbf{e}_{c}^{\ulcorner i,j} \rVert^{2} ~, 
\end{equation}
where $\mathbf{e}_{c}$ denotes the consensus error term which is responsible to enforce the constraints in Eq. \eqref{eq:pose_consensus_constraint}.
As we perform consensus based on variables in $\mathbb{SO}3$, special care needs to be taken while handling the consensus terms in Eq. \eqref{eq:arock_min}.
To do so, we propose to perform the consensus on the tangent space of a fixed reference rotation, leading to a modified version of Eq. \eqref{eq:pose_consensus_constraint} in terms of the rotation:
\begin{equation}
\label{eq:tangent_pose_consensus}
\delta \mathbf{q}_{A,A}^{i} = \delta \mathbf{q}_{A,B}^{i}, \quad \delta \mathbf{q}_{B,A}^{i} = \delta \mathbf{q}_{B,B}^{i} \quad \forall i \in [1, N - l]~,
\end{equation}
where for example $\delta \mathbf{q}_{A,A}^{i}$ is defined via the following relation
\begin{equation}
\label{eq:definition_tangent_vector}
\mathbf{q}_{MS_{A,A}}^{i} = \mathbf{q}_{A,ref}^{i} \boxplus \delta \mathbf{q}_{A,A}^{i}.
\end{equation}
The reference rotation $\mathbf{q}_{A,ref}^{i}$ is fixed and is chosen to be the estimated rotation after the pose tracking.
Following this definition, we can write down the terms $\mathbf{e}_{c}^{i,j}$ as 
\begin{align}
\mathbf{e}_{c,q}^{i,j} &= \left( \frac{1}{2 \gamma_{q}}\mathbf{z}_{i,q}  + \delta \mathbf{q}_{i,j}\right) \sqrt{2 \gamma_{q}}, \text{ and} \label{eq:rotation_consensus_error} \\
\mathbf{e}_{c,p}^{i,j} &= \left( \frac{1}{2 \gamma_{p}}\mathbf{z}_{i,p}  + \mathbf{p}_{MS_{i,j}}\right) \sqrt{2 \gamma_{p}} ~. \label{eq:translation_consensus_error}
\end{align}
Note that inserting these into Eq. \eqref{eq:distributed_objective} is equivalent to the formulation in Eq. \eqref{eq:arock_min}, however, as in our implementation we use the Ceres \cite{ceres} library to solve the minimization, by using the re-arranged notation the consensus-errors can be directly inserted. 
The update of the dual variables as in Eq. \eqref{eq:arock_left}, \eqref{eq:arock_right} for the rotational part of the poses is again performed in the tangent space of the associated reference rotations
\begin{equation}
\label{eq:dual_update_rotation}
\mathbf{z}_{\mathbf{q}_{i,j}}^{k+1} = \mathbf{z}_{\mathbf{q}_{i,j}} - \eta \left( (\mathbf{z}_{\mathbf{q}_{i,j}}^{k} + \hat{\mathbf{z}}_{\mathbf{q}_{i,j}}) + \gamma \delta \mathbf{q}_{i}^{k+1} \right) ~.
\end{equation}
The translational part is update as defined in Eq. \eqref{eq:arock_left}, \eqref{eq:arock_right}.
The obtained dual variables $\mathbf{z}_{i,j}^{k}$ are then passed to the communication module \refsec{ssec:communication}, which transmits the information to the other agent.
As the \gls{admm} scheme used is iterative by design and is employed in a sliding window fashion, we continuously run the optimization.
Similarly as proposed in \cite{Zhang:etal:PAMI2018}, we only execute a limited number of iterations in the minimization step.
At the end of every minimization step of Eq. \eqref{eq:distributed_objective}, we shift the sliding window and remove the \glspl{kf} that fall outside the window and remove \glspl{mp} that have no observations within the shifted window.

\subsection{Communication}
\label{ssec:communication}
The communication module of the proposed pipeline is responsible to communicate newly created data (i.e. \glspl{kf}) as well as for exchanging updates on the dual variables over a wireless network.
In the proposed system we employ ROS \cite{ros} to serialize/de-serialize data and communicate it between the agents.
Every time a new \gls{kf} gets inserted in the window, this \gls{kf} is passed to the Communication module.
As sending the full data of this \gls{kf}, e.g. image information, would lead to tremendous network traffic, we summarize the necessary information into a message containing the following information:
\begin{itemize}
\item \gls{kf} timestamp and \gls{kf} ID information
\item 2D keypoint locations and extraction octave information
\item Keypoint descriptors
\item Pose $\mathbf{T}_{MS}$ after tracking
\item Covariance $\pmb{\Sigma}_{\mathbf{T}_{MS}}$ of the tracked pose.
\end{itemize}
Note that the sent pose is not only used as an initial guess, but it also defines the reference rotation used in the pose consensus.\\
In order to update the consensus error terms, the dual variables associated to the \glspl{kf} within the optimization window get exchanged after every update step in Eq. \eqref{eq:arock_left}, \eqref{eq:arock_right}.
As the dual variables are uniquely assigned to a \gls{kf} pose, we use the identifier of the associated \gls{kf} in order to correctly assign the dual variables.
In summary, for every performed update on the dual variables, we stack the following information into a message and transmit it to the neighboring agent:
\begin{itemize}
\item Identifier of the sender (i.e. the agent ID)
\item Vector of dual variables (in 6D)
\item Vector of associated \gls{kf} IDs
\end{itemize} 
Note that the latest received dual variables constitute the $\hat{\mathbf{z}}$ in Eq. \eqref{eq:arock_min}-\eqref{eq:arock_right}.

\subsection{Initialization}
\label{ssec:initialization}
While the previous sections describe the normal mode of operation for the system, an initialization procedure is required in order to build up the initial conditions.
As this initialization phase is only active for a limited time, we perform all the necessary computations on a single agent in order to increase the ease the implementation. \\
To bootstrap the map, based on the timestamp we select the two frames from the agents, which are closest together in time and perform brute force descriptor matching to find 2D-2D correspondences.
After a 2D-2D RANSAC outlier rejection, the inlier correspondences are used to compute the relative pose up to scale between the two frames.
In order to remove the scale ambiguity, we scale the obtained baseline between the initial frames using the closest available \gls{uwb} measurement and attempt to triangulate the inlier correspondences to generate \glspl{mp}.
Every subsequent frame (both native and non-native), gets matched and aligned against the existing \glspl{mp} using a zero-velocity motion model.
After the alignment, we attempt the triangulation of new \glspl{mp} using the newest frame-pair and vision-based bundle adjustment is performed optimizing the frame poses and the \glspl{mp}.
In the case, where we are unable to align a frame, i.e. due to lack of sufficient correspondences, the initial Map is reset and we start with another initial pair again.
This process is repeated until the number of frames is sufficient to create at least 4 \glspl{kf}. \\
Once the minimal number of \glspl{kf} is reached, we aim to initialize the \gls{imu} related variables for the tracking, namely $\mathbf{q}_{WM}, {}_{S}\mathbf{v}$ and $\mathbf{b}_{\omega}$.
For this, we utilize use a method inspired by the \gls{imu}-initialization proposed by \cite{Qin:etal:TRO2018}, which first estimates the gyroscope biases $\mathbf{b}_{\omega}$ and then solves for velocity states and gravity direction in a least-squares fashion.
As we perform the initialization for two trajectories, we jointly solve for the velocities of both agents, but only a single, shared gravity direction.
Here, in contrast to \cite{Qin:etal:TRO2018}, as we utilize the \gls{uwb} measurements, we are able to avoid including the scale as an estimation parameter.
After successful initialization, the agent responsible for the computations sends the all of its native \glspl{kf} along with the \glspl{mp} and the \gls{imu} states to the other agent, and from then on the system enters the normal mode of operation as outlined in the previous sections.

\section{Experimental Evaluation}\label{sec:experiments}
\subsection{Run-time efficiency of combined Z-spline interpolation}
\label{ssec:z_spline_runtime}
\begin{figure}[t]
    \centering
    \vspace{2mm}
    ~ 
    \begin{subfigure}[b]{0.48\columnwidth}
        \includegraphics[trim={1.8cm 0 1.8cm 0}, clip, width=1.0\columnwidth]{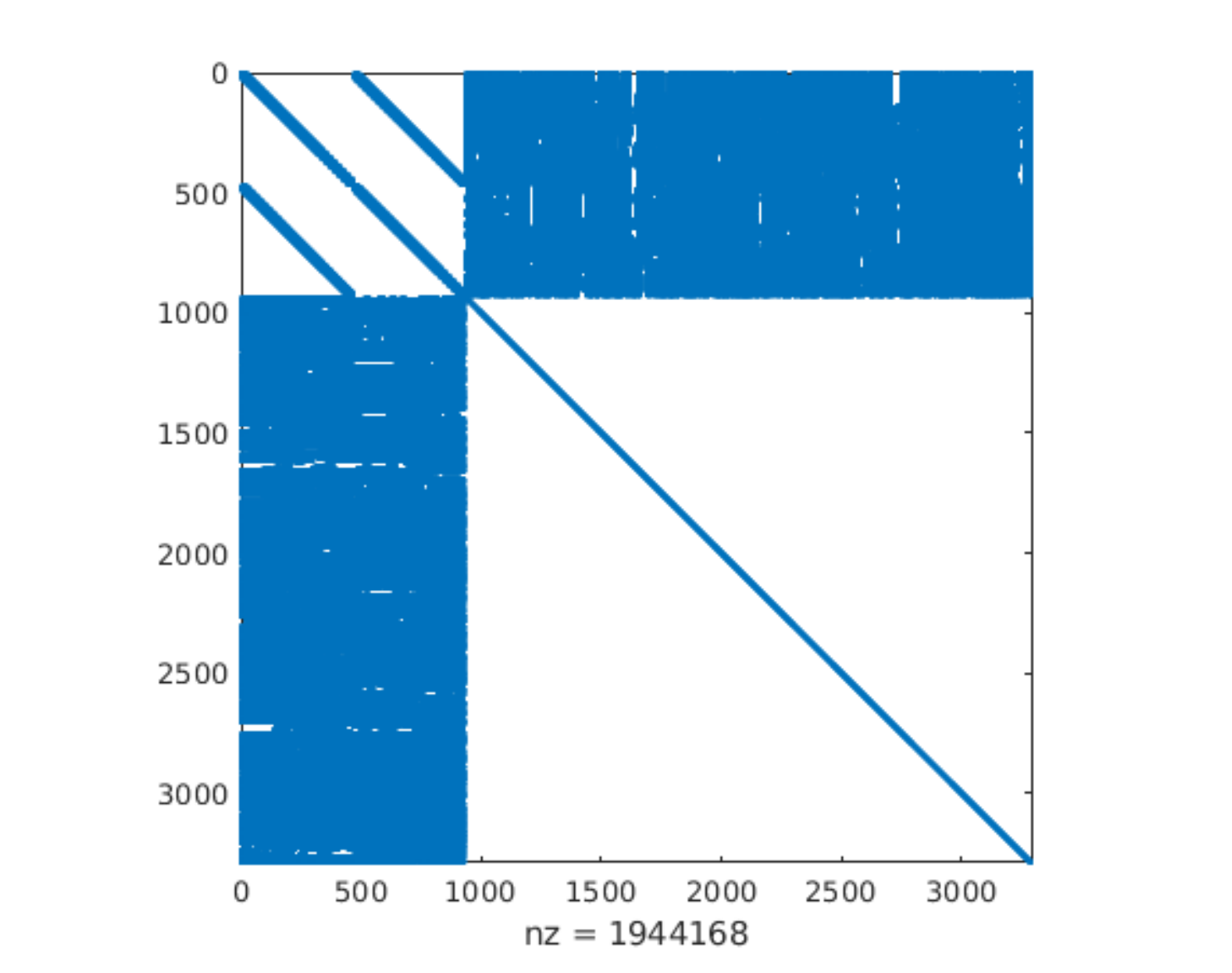}
        \caption{B-spline}
        \label{fig:basis_functions_plot}
    \end{subfigure}
	\begin{subfigure}[b]{0.48\columnwidth}
        \includegraphics[trim={1.8cm 0 1.8cm 0}, clip, width=1.0\columnwidth]{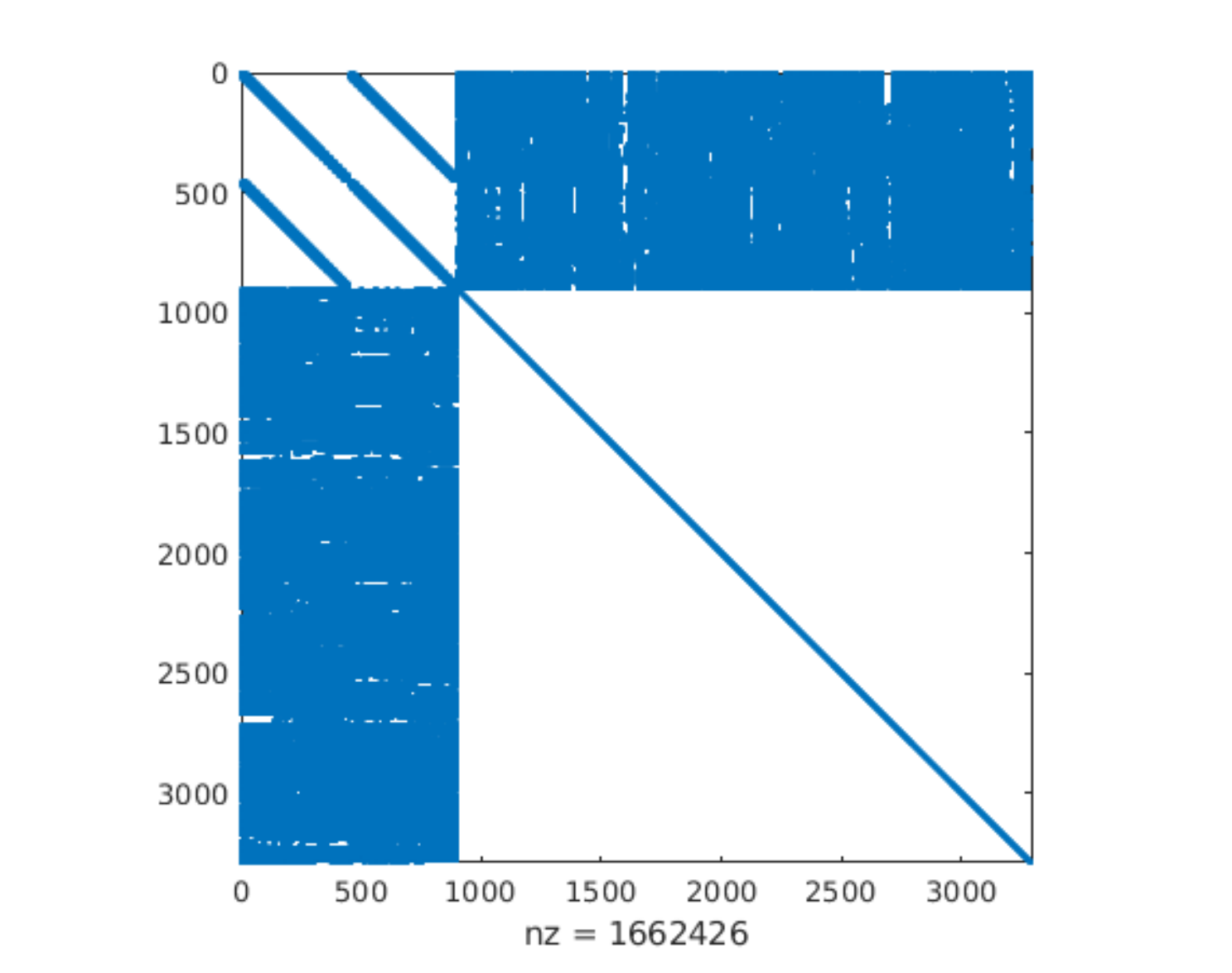}
        \caption{Combined Z-spline }
        \label{fig:cum_basis_functions_plot}
    \end{subfigure}
    \caption{Comparison between the resulting Hessian matrix for the same problem when using standard B-spline based interpolation (a) versus the usage of the proposed combined Z-spline based KF-interpolation (b). While the structure is very similar in both cases, the non-zero elements (``nz'') are reduced by $15\%$ in (b)}
\label{fig:hessian_structure}
\end{figure}

In this section, we investigate the influence of the trajectory representation proposed here using a combination of \gls{kf} poses and a Z-spline based interpolation along these poses on the run-time efficiency. 
As, computationally, the most expensive part in the proposed system is the minimization of Eq. \eqref{eq:distributed_objective}, reducing the execution time of this minimization permits the execution of more distributed optimization iterations, which in return potentially boosts the accuracy of the estimates.\\
As the minimization in  Eq. \eqref{eq:distributed_objective} is performed using a second order method, the iteration time is, to a large extent, determined by the time required to solve the resulting normal equations, which in turn is highly depending on the sparsity of the resulting Hessian matrix. 
As a result, we investigate the influence of the proposed trajectory representation on the sparsity of the resulting Hessian matrix.
For the relative distance error terms in Eq. \eqref{eq:relative_distance_residual} there is no structural difference between the combined Z-spline based interpolation and the standard B-spline interpolation.
On the other hand, for the reprojection errors, the Jacobian structure changes significantly.
While computing the error using the traditional B-spline based interpolation four base poses have a non-zero contribution in the Jacobian, utilizing the \glspl{kf} as base poses for a Z-spline based interpolation, one can directly compute the reprojection error using the \gls{kf}-pose.
As a result, the number of non-zero elements in the underlying Jacobian matrix is roughly $1/4$ (as the number of residuals is dominated by the reprojection errors), compared to the standard B-spline interpolation.\\
Using a sample over a trajectory length of 8 seconds with a \gls{kf} interval of $0.15$ seconds and roughly 2000 \glspl{mp} we build up the normal equation for the same problem, both using B-spline interpolation as well as the proposed interpolation scheme and extract the resulting Hessian matrices.
In \reffig{fig:hessian_structure} the corresponding Hessians with the number of non-zero elements are shown.
While the structure itself is similar in both cases, one can see that the proposed Z-spline based formulation improves the sparsity of the resulting Hessian matrix by around 15\%.
The sparsity remains after performing factorization, indicating that the proposed interpolation scheme indeed reduces the computational complexity in the solving step.

\subsection{Runtime and Bandwidth Evaluation}
\label{ssec:runtime_bandwidth}
As real-time capabilities and data sharing rates are crucial elements for the applicability of a distributed system, in this section we analyze the run-time of the main elements of the proposed system and investigate the bandwidth requirements for the necessary data exchange.
In order to evaluate the system under realistic conditions, all the experiments are performed using two Intel Core i7-8550U computers communicating with each other via a wireless TP-Link AC1750 router configured to use the $2.4$GHz interface.
\subsubsection{Runtime Evaluation}
\label{sssec:runtime_evaluation}
As the proposed system runs multiple threads, the decisive element for the real-time capability of the overall system is the tracking-thread, which needs to process every frame.
Nonetheless, the timings of the other parts, i.e. mapping and optimization, are important for the overall performance of the system and thus are measured.
For example, a slow optimization step will result in fewer \gls{admm}-iterations, possibly leading to bigger inconsistencies and inaccuracies.
In \reftab{tab:timings}, the timings for the main tracking- and mapping processes are reported.
On the tracking side, the most expensive operation is the extraction of the BRISK features with approximately $8$ms per frame, while the matching-step and EKF estimation roughly take up to $2$ms per frame.
The total latency describes the effective time it takes from the reception of the image until the pose estimate is computed, which takes on average $10$ms.
Even in the worst case recorded, the latency stays within a $25$ms budget, which corresponds to twice real-time.
The timings for mapping are dominated by the correspondence search, while the time for uncertainty maintenance and triangulation is almost neglectable. 
Note that the EKF based uncertainty maintenance of the \glspl{mp} gets called for every \gls{kf}, which in our case corresponds to every third frame, whereas the matching and triangulation processes only get called if new \glspl{mp} need to be inserted.
The optimization loop, which includes communicating and updating the dual variables, takes approximately $77$ms on average with a standard deviation of $11.5$ms. 
With a \gls{kf}-interval of $150$ms, this means on average per \gls{kf}, two \gls{admm}-iterations are performed.

\begin{table}[]
\begin{tabular}{|c|l|c|c|c|}
\cline{3-5}
\multicolumn{2}{c|}{}  & \bf{mean $\pm \pmb{2\sigma}$ {[}ms{]}} & \bf{max {[}ms{]}} & \bf{rate {[}Hz{]}}         \\ \hline                  
\multirow{4}{*}{\rotatebox[origin=c]{90}{ \bf{Tracking} }} & BRISK Extraction & $7.9\pm 3.4$ & $20.8$                    & \multirow{4}{*}{$20$} \\
[0.7ex] 
                  & Matching        & $0.5\pm 0.4$         & $8.7$         &                       \\
[0.7ex] 
                  & Tracking EKF    & $1.4 \pm 1.3 $         & $12.0$        &                       \\
[0.7ex]
                  & Total Latency   & $10.3 \pm 4.2$         & $25.0$        &                       \\ \hline               
\multirow{3}{*}{\rotatebox[origin=c]{90}{\bf{Mapping}}} & Uncertainty EKF & $0.2 \pm 0.1$         & $1.4$         & $6.7$                  \\
[0.7ex] 
                  & BF-Matching     & $8.3 \pm 8.5$         & $26.1$        & $1.6$ \\
[0.7ex] 
                  & Triangulation    & $0.1 \pm 0.5$ & $1.8$ &          $1.6$      \\ \hline 
\end{tabular}
\caption{Runtime evaluation of the different parts of the tracking and mapping, recording the mean $\pm$ twice the standard deviation of the respective measurement, and the rates of the corresponding operations performed on average.}
\label{tab:timings}
\end{table}

\subsubsection{Bandwidth Usage}
\label{sssec:bandwidth_usage}
\begin{figure}[t]
    \centering
    \vspace{2mm}
    ~ 
        \includegraphics[trim={0 0 0 0}, clip, width=1.0\columnwidth]{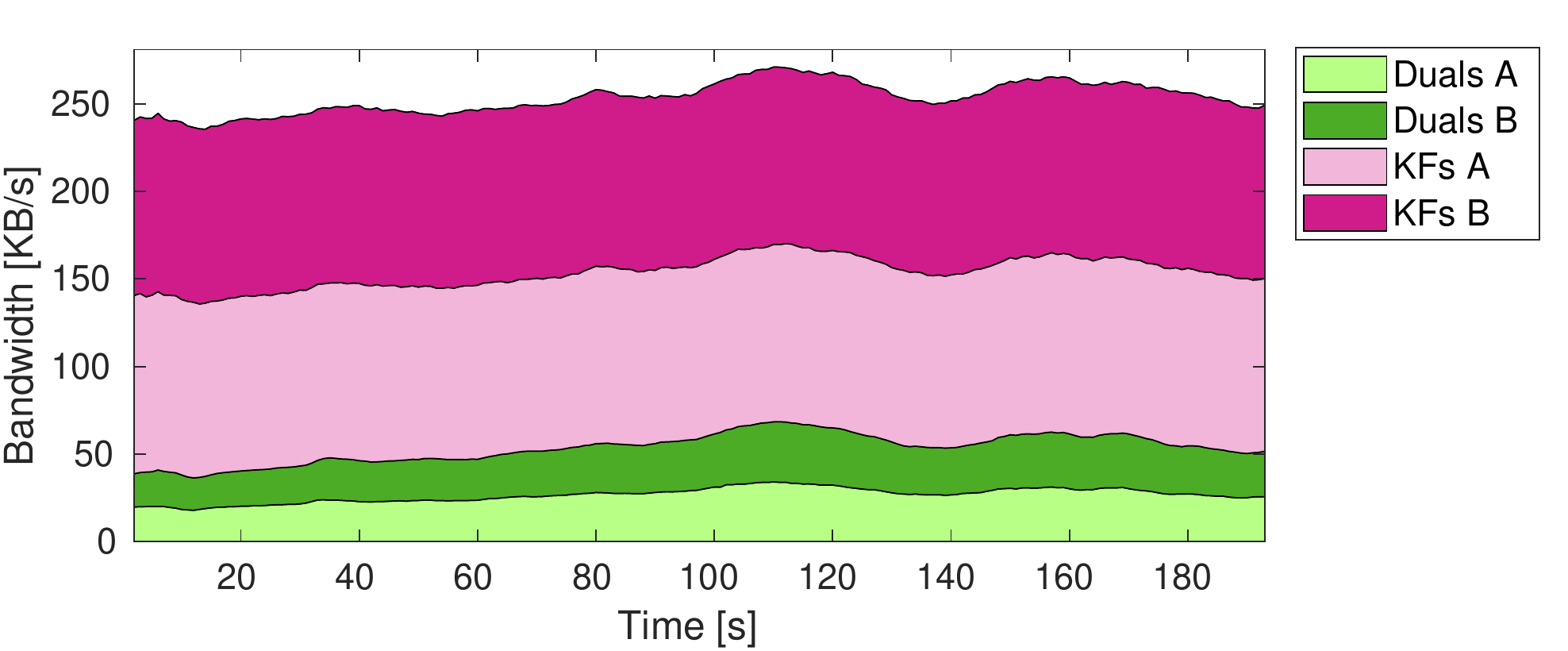}
    \caption{Bandwidth usage over time for the different messages and agents during a single experiment. The total bandwidth requirement is approximately $250$KB/s, which is easily possible using e.g. standard WiFi communication.}
\label{fig:bandwidth}
\end{figure}
In the proposed system, the data exchanged consists of \gls{kf}-data and the dual variables exchanged in the \gls{admm} scheme.
As we use ROS for the communication, the built-in functionality for monitoring bandwidth usage was used to obtain the data presented in this section.
\reffig{fig:bandwidth} shows an example of the required bandwidth for one of the datasets used in \refsec{ssec:comparison_vio}. 
As it can be observed, the total bandwidth requirement remains generally constant around $250$KB/s, whereas the majority of the exchanged data is contained in the \glspl{kf} with approximately $100$KB/s per agent, while the exchange of the dual variables generates about $25$KB/s of network traffic per agent.
Note that while on the basis of the fixed optimization window of $5$s, the package size for the exchange of the dual variables is constant, the rate of the exchange is coupled with the frequency of the optimization-loop, which is subject to fluctuations.
A summary of the bandwidth usage over a larger set of experiments is provided in \reftab{tab:bandwidth}.
As it can be observed, the overall bandwidth usage is not subject to large fluctuations and on average is below $250$KB/s, which is easily feasible with a standard WiFi module (e.g. IEEE 802.11g standard).

\begin{table}[h]
\centering
\begin{tabular}{l | c | c}
               & \bf{mean $\pm \pmb{2\sigma}$ {[}KB/s{]}} & \bf{max {[}KB/s{]}} \\
               \hline 
               \hline
Dual Variables & $48.6\pm 14.0$            & $68.5$           \\ \hline
Keyframes      & $197.3 \pm 8.8$           & $214.5$         \\ \hline
Total          & $245.8 \pm 19.1$           & $270.9$         
\end{tabular}
\caption{Bandwidth usage of the different message types summed together for both agents, recording the mean $\pm$ twice the standard deviation for the corresponding measurements.}
\label{tab:bandwidth}
\end{table}

\subsection{Photo-Realistic Synthetic Datasets}
\label{ssec:synthetic_datasets}
\begin{figure*}[t]
    \centering
    \vspace{2mm}
    ~ 
    \begin{subfigure}[b]{0.24\textwidth}
        \includegraphics[trim={0 0 0 0}, clip, width=1.0\textwidth]{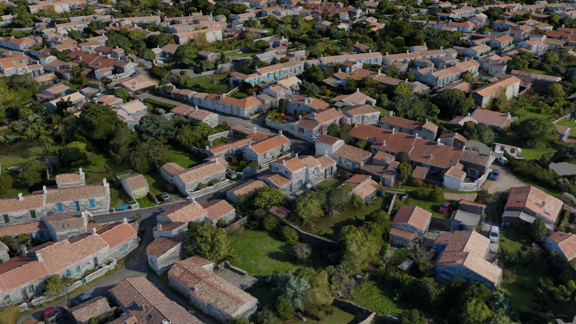}
        \caption{Scene 1}
        \label{fig:scene_medicon}
    \end{subfigure}
	\begin{subfigure}[b]{0.24\textwidth}
        \includegraphics[trim={0 0 0 0}, clip, width=1.0\textwidth]{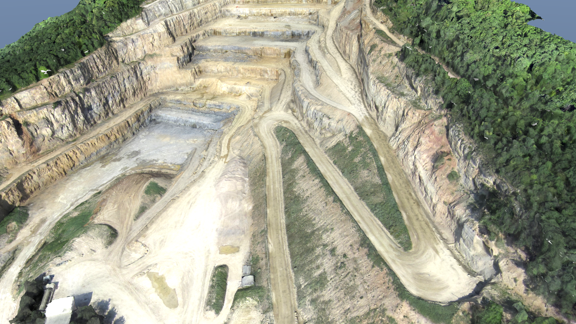}
        \caption{Scene 2}
        \label{fig:scene_medicon_mountain}
    \end{subfigure}
    \begin{subfigure}[b]{0.24\textwidth}
        \includegraphics[trim={0 0 0 0}, clip, width=1.0\textwidth]{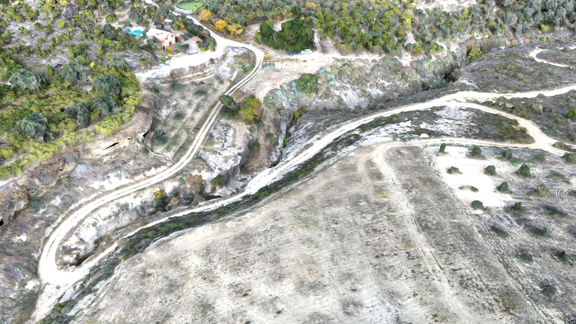}
        \caption{Scene 3}
        \label{fig:scene_city}
    \end{subfigure}
    \begin{subfigure}[b]{0.24\textwidth}
        \includegraphics[trim={0 0 0 0}, clip, width=1.0\textwidth]{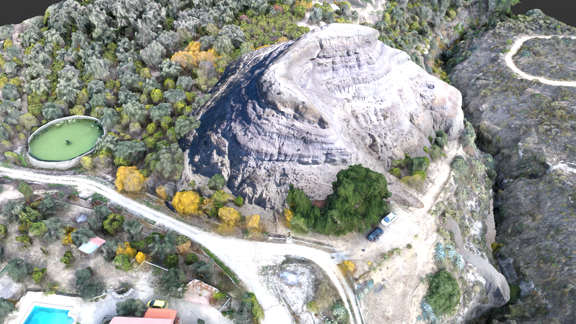}
        \caption{Scene 4}
        \label{fig:scene_quarry}
    \end{subfigure}
    \caption{Snapshots of the four photo-realistic scenes used to render experimental data for the evaluation of the proposed system}
\label{fig:photorealistic_models}
\end{figure*}

Performing outdoor experiments with two \glspl{uav} flying relatively close to each other at high altitudes is extremely challenging as at that altitude the pilot has very limited visual understanding of the motion of the \glspl{uav}, posing a significant risk of losing control of the aircraft.
Furthermore, obtaining ground-truth of the robots' poses allowing a quantitative evaluation of the proposed framework on real data is problematic due to fluctuations in the accuracy of GPS measurements and the challenging estimation of the orientation.
As a result, inspired by the idea in \cite{Teixeira:etal:RAL2020}, we create synthetic photorealistic datasets from real images to test and evaluate the proposed approach.
Analogously to \cite{Teixeira:etal:RAL2020}, we simulate both UAVs' dynamics using the RotorS Gazebo simulator \cite{Furrer:etal:ROS2016} and utilize the Blender render engine to generate the associated image data from 3D models obtained by photogrammetric reconstruction\footnote{\url{https://github.com/VIS4ROB-lab/visensor_simulator}}. 
The visual-inertial sensor data was simulated to mimic the data-stream obtained from a sensor as in \cite{Nikolic:etal:ICRA2014}, consisting of a global-shutter grayscale image-stream, time-synchronized with the \gls{imu} data.
The images are rendered with a resolution of $480\times752$ at $20$Hz and the \gls{imu} data has a rate of $200$Hz.
The \gls{uwb}-distance measurements are simulated by computing the ground-truth relative distance between the \glspl{uav} and disturb it with gaussian noise.
The \gls{uwb}-data is simulated at $60$Hz with a noise level of $0.1$m standard deviation.
The simulated \glspl{uav} have a maximal diameter (from tip-to-tip) of $0.85$m and during all experiments, we set the smallest allowed baseline to be $1.0$m.\\
In order to control the relative pose between the two \glspl{uav}, we employ the control strategy as outlined in \refsec{ssec:active_baseline_control}, while the estimated scene depth is obtained by computing the median depth from a rendered depth image.
We use four different scenes, as shown in \reffig{fig:photorealistic_models}, and rendered multiple trajectories for each scene. \\
\textbf{Scene 1}: A suburban housing settlement consisting of smaller houses with gardens, small parks and connecting streets spanning an area of approximately $450\text{m} \times 450\text{m}$. The, from high altitudes, plane-like structure along with the well textured scene, results in well suited imagery for vision based methods.\\
\textbf{Scene 2}: A mine with different levels of erosion, small trails and some parts with steep flanks, resulting in abrupt changes in the scene depth of up to $35\text{m}$. The steep flanks in combination with the, at times, uniform texture prohibit lengthy and uniform feature association.\\
\textbf{Scene 3}: Mediterranean countryside with sections containing different kinds of vegetation as well as some small canyons and trails offering a mixture of texture-rich and low-textured areas along with some depth variation across the canyons.\\
\textbf{Scene 4}: Same structure as Scene 3 but with an artificially added hill structure of about $40\text{m}$ height in order to increase the depth variations in the scene.

\subsection{Comparison to Visual-Inertial SLAM at higher altitudes}
\label{ssec:comparison_vio}
\begin{figure}[t]
    \centering
    \vspace{2mm}
    ~ 
    \begin{subfigure}[b]{0.85\columnwidth}
        \includegraphics[trim={0cm 0 0cm 0}, clip, width=1.0\columnwidth]{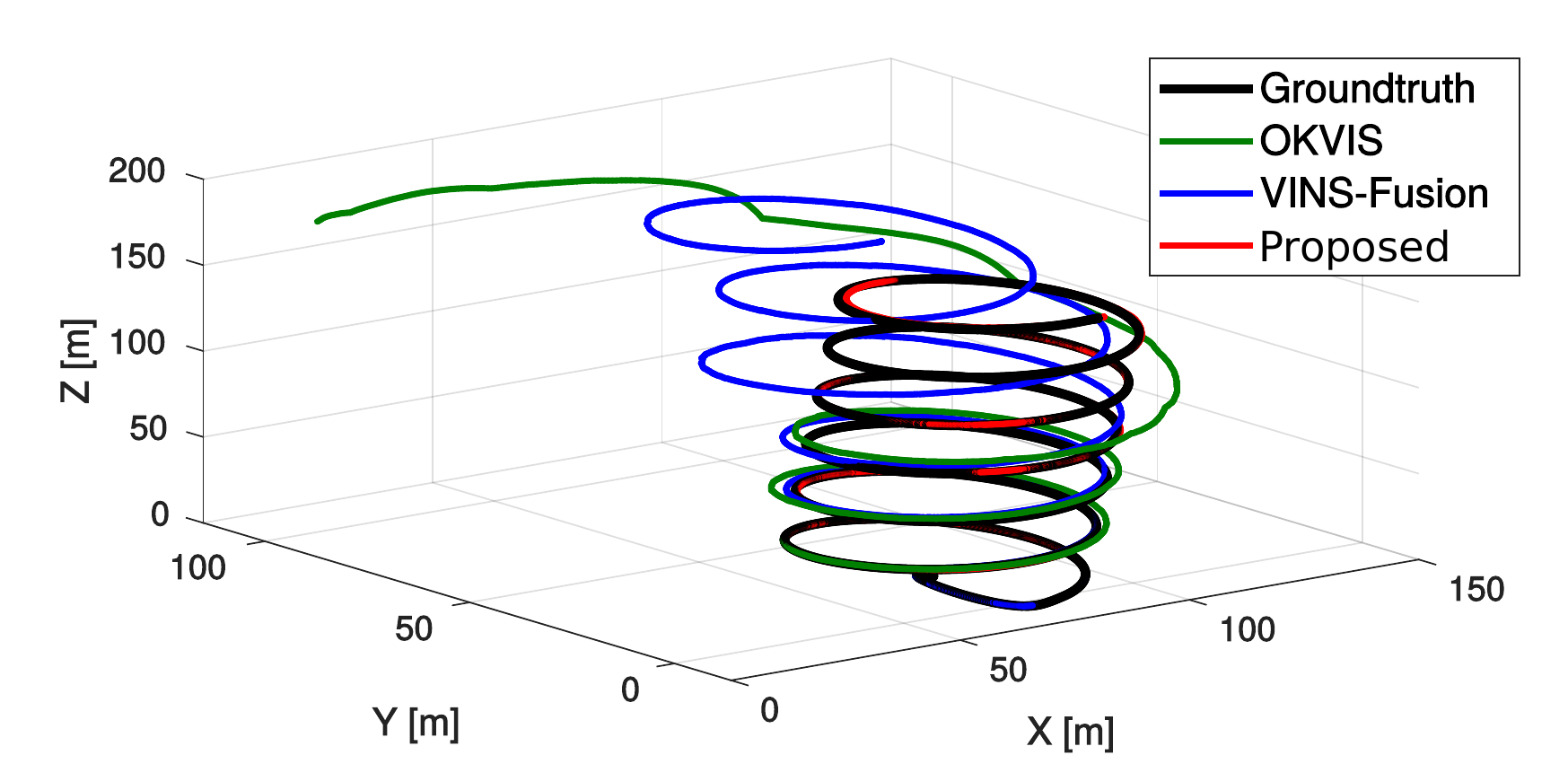}
        \caption{Example run for a spiral trajectory on Scene 1}
        \label{fig:spiral}
    \end{subfigure}
	\begin{subfigure}[b]{1.0\columnwidth}
        \includegraphics[trim={0cm 0 0cm 0}, clip, width=1.0\columnwidth]{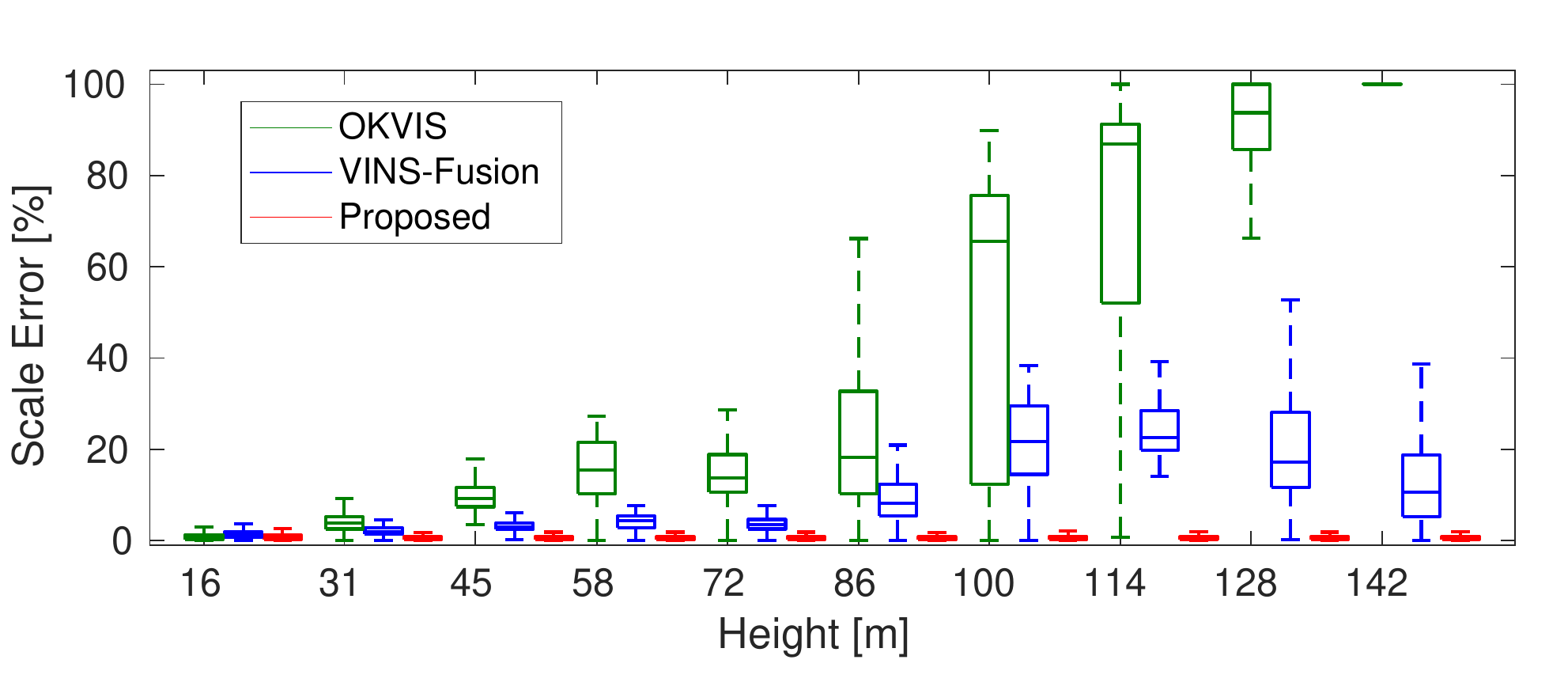}
        \caption{Scale error for different altitudes}
        \label{fig:scale_error}
    \end{subfigure}
    \caption{Evaluation results for spiral trajectories with increasing altitudes. In (a) is the result of a single run, where the estimates of both stereo \glspl{vio} start to diverge with increasing altitude, while (b) summarizes the scale errors over all performed spiral experiments, where it is evident that the effectiveness of the imposed scale constraints becomes an issue for (stereo-) \gls{vio} methods with increasing altitude.}
\label{fig:spiral_experiment}
\end{figure}

\begin{figure}[t]
    \centering
    \vspace{2mm}
    ~ 
        \includegraphics[trim={0 0 0 0}, clip, width=1.0\columnwidth]{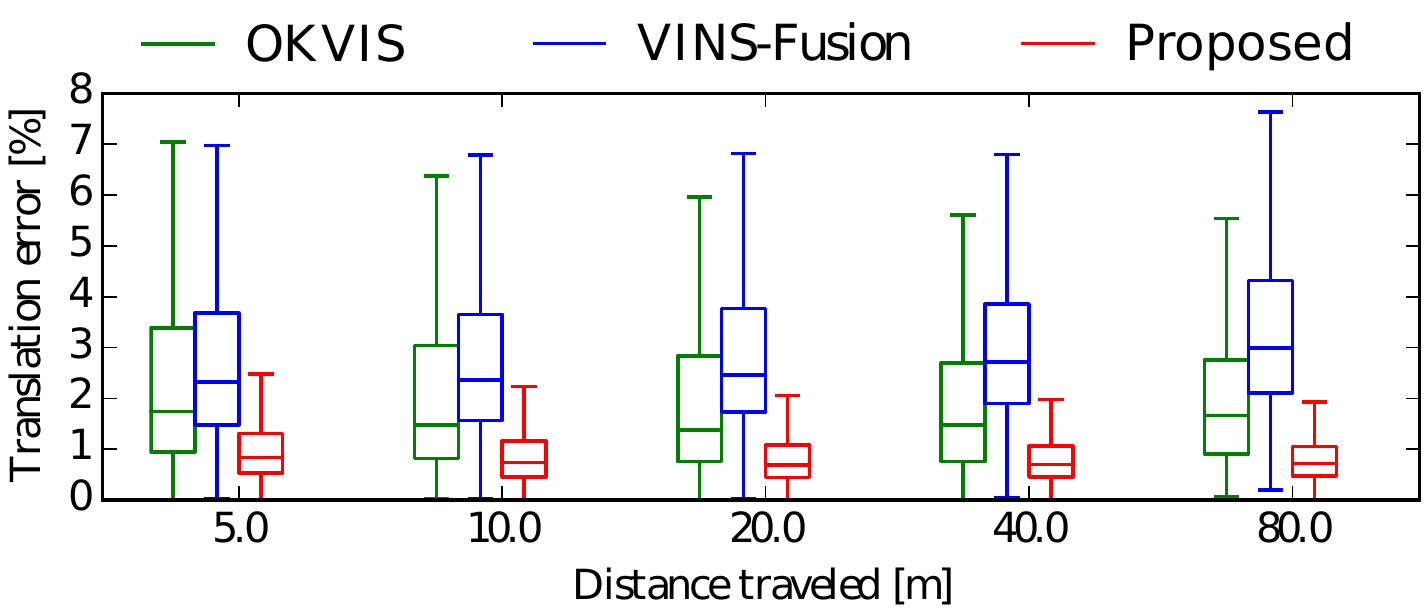}
    \caption{Comparison of the relative odometry error between state-of-the-art stereo \gls{vio} methods and the proposed approach. The reported errors are obtained by computing the error statistics over all datasets over 3 runs.}
\label{fig:vio_comparison}
\end{figure}

In this section, we provide a comparison of the proposed approach against the two highest performing state-of-the-art stereo \gls{vio} methods that are publicly available, namely VINS-Fusion \cite{Qin:etal:TRO2018} and OKVIS \cite{Leutenegger:etal:IJRR2015}, to test their accuracy of estimates at increasing scene depths.
To illustrate this effect, we generated spiral-shaped trajectories over Scenes 1-3 with gradually increasing altitude.
The trajectories have a radius of 25 meters and increase their height with a rate of 25 meters per turn up to a height of approximately 160 meters. 
The fixed stereo camera baseline for VINS-Fusion and OKVIS was chosen to be $0.22$m, which was chosen to be twice the size of the sensor proposed in \cite{Nikolic:etal:ICRA2014}. \\
An example of such a trajectory along with the aligned estimates of both our system and the stereo-inertial estimators is shown in \reffig{fig:spiral}.
Initially, all the estimates are close to the ground-truth, however, with increasing height, both estimates of VINS-Fusion as well as OKVIS start to diverge with increasing altitude, while the estimate of the proposed system remains close to the ground-truth trajectory.
As it can be observed, the trajectories of the stereo \glspl{vio} are mainly fluctuating in scale and less in the shape of the trajectory itself. 
In order to evaluate the scale uncertainty, we perform an evaluation in a similar fashion to a relative error evaluation as described in \cite{Zhang:etal:IROS2018}.
We select a sub-trajectory of a given length (here 100 frames), align it to the ground-truth with a similarity transform as computed using the approach in \cite{Umeyama:PAMI1991} and record the scale. 
The selected sub-trajectory is then shifted by 5 frames and the alignment is repeated until the end of the trajectory is reached, resulting to statistics of the scale error.
%
Using the altitude as an approximation of the scene depth, in \reffig{fig:scale_error}, the scale-error statistics with respect to the altitude are reported.
As it can be observed, with increasing altitude, the scale errors quickly become significant resulting in a large uncertainty of the estimate. 
While VINS-Fusion generally shows a slower increase of the scale error than OKVIS, both \gls{vio} methods suffer from the same tendency with increased scene depth, whereas the proposed method exhibits a constant scale uncertainty over the full altitude range. \\
In order to allow for a quantitative comparison between our proposed approach to the two state-of-the-art stereo \glspl{vio}, different trajectories at altitudes ranging between $15-35$ meters are rendered.
In particular two different trajectories are created for Scenes 1-3 summing up to a total trajectory length of $3.4$km.
For the proposed method, the baseline between the two agents is chosen to be roughly $2$m, while this is allowed to fluctuate slightly without considering the particular scene depth.
The resulting relative comparison between the relative odometry errors of the stereo \gls{vio} methods and the proposed method is shown in \reffig{fig:vio_comparison}.
As it can be observed, compared to the existing \gls{vio} methods, the proposed system has both a lower median error as well as smaller fluctuations in the statistics, as expected, due to the fact that the selected baseline should exhibit more favorable behavior for the overall scene depth.
Between the two existing stereo \glspl{vio}, OKVIS performs slightly better than VINS-Fusion on the evaluated datasets, whereas to a large extent the increased error of the latter can be traced back to an increased yaw-drift. 
Note that the evaluated datasets are closer to the intended use-cases of the proposed system, i.e. scene depths above $10$m, than the \glspl{vio}, which are generally designed and evaluated on datasets closer to the ground or indoors.
Hence, we do not claim to outperform these systems in the general case, however, we can show the advantage of the proposed approach already at flying altitudes significantly below the heights where the stereo \gls{vio} systems start to fail.

\subsection{Active Baseline Control}
\label{ssec:active_baseline_control}
\begin{figure}[t]
    \centering
    \vspace{2mm}
    ~ 
        \includegraphics[trim={0 0 0 0}, clip, width=1.0\columnwidth]{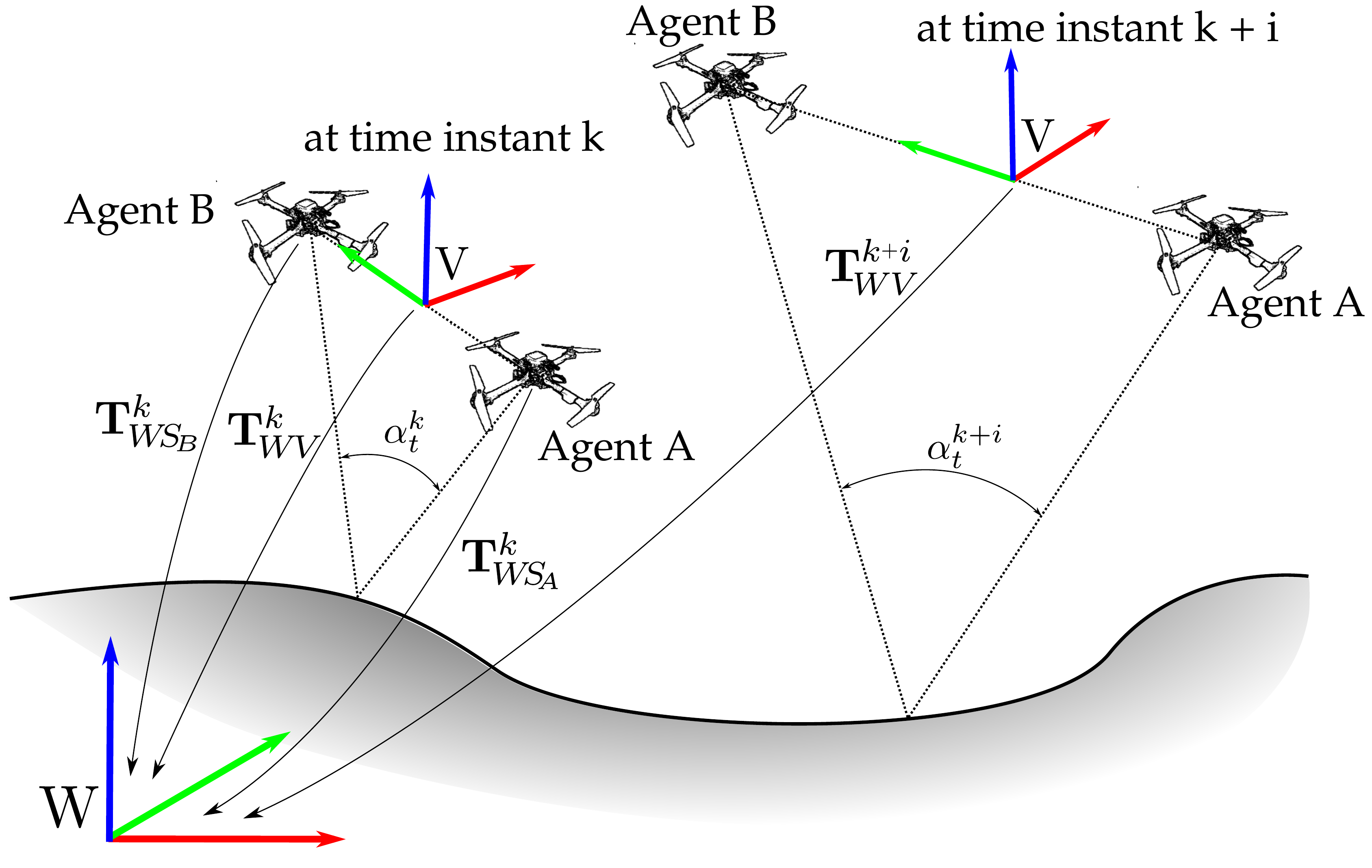}
    \caption{Illustration of the virtual stereo setup with two agents. Using the desired location of the virtual stereo center $V$ and a target triangulation angle $\alpha_{t}$, the required agent poses are computed.}
\label{fig:baseline_controller}
\end{figure}

As the proposed collaborative system estimates the poses of the agents in a common reference frame, we are able to control the relative distance between the \gls{uav} agents. 
For this purpose, we designed a high-level controller with the goal of controlling the agents' baseline in order to form a virtual stereo camera, as illustrated in \reffig{fig:baseline_controller}.
For simplicity, we assume that the monocular cameras are mounted at an identical viewing angle onboard each agent and that aligning the agents along their X-axes results to a valid stereo configuration.
Note that with a few additional computations, this approach can be adapted to a more generic setup, such as having different mounting angles of the cameras on the \glspl{uav},  as well.
The relative translation between the cameras in the virtual stereo setup is given by the average of the two agents' poses:
\begin{equation}
\label{eq:virtual_stereo_center}
\mathbf{p}_{WV}^{k} = \frac{1}{2}\left( \mathbf{p}_{WS_{A}}^{k} + \mathbf{p}_{WS_{B}}^{k} \right) ~.
\end{equation}
In order to obtain the yaw angle, we first project the relative baseline onto the X-Y plane of the inertial frame
\begin{equation}
\label{eq:virtual_stereo_proj}
\mathbf{p}_{proj} = \begin{bmatrix}
1 & 0 & 0 \\
0 & 1 & 0
\end{bmatrix} \left( \mathbf{p}_{WS_{B}}^{k} - \mathbf{p}_{WS_{A}}^{k} \right)
\end{equation}
and from that we compute the resulting orientation as:
\begin{equation}
\label{eq:vritual_stereo_yaw}
\psi_{WV} = \mathrm{atan2}\left( \mathbf{p}_{proj,y}, \mathbf{p}_{proj, x} \right) + \pi / 2 ~,
\end{equation}
where $\mathbf{p}_{proj,x}$ corresponds to the first and $\mathbf{p}_{proj,y}$ to the second entry of $\mathbf{p}_{proj}$. \\
For a given $4$-DoF target pose ($x_{t}, y_{t}, z_{t}, \psi_{t}$) and the scene depth $d_{s}$, suitable poses for the agents can be computed by reformulating Equations \eqref{eq:virtual_stereo_proj}, \eqref{eq:vritual_stereo_yaw}.
Using the desired triangulation angle $\alpha_{t}$ (here set to $10^{\circ}$) and $d_{s}$, the resulting baseline of the virtual stereo camera is given by
\begin{equation}
\label{eq:virtual_stereo_baseline}
b_{V} = 2 \cdot \mathrm{tan}(\alpha_{t} / 2) ~.
\end{equation}
Hence, the resulting agents' target translations are given by
\begin{equation}
\label{eq:virtual_stereo_uav_translations}
\mathbf{p}_{WS_{A/B}} = \begin{bmatrix}
x_{t} \\ y_{t} \\ z_{t}
\end{bmatrix} + \begin{bmatrix}
\cos(\psi_{t}) & \sin(\psi_{t}) & 0 \\
\sin(\psi_{t}) & cos(\psi_{t})) & 0 \\
0 & 0 & 1
\end{bmatrix} \begin{bmatrix}
0 \\
\mp b_{V} / 2 \\
0
\end{bmatrix} ~,
\end{equation}
where the minus sign corresponds to agent A and the plus sign to agent B, assuming that agent A acts as the right camera in the virtual stereo setup.
The target yaw angles of the agents are set to the target yaw angle of the virtual stereo camera.
The computed target poses for both agents are fed to the MPC-based position controller running on each agent \cite{Kamel:etal:ROS2017}.
Note that in order to limit the speed of the response, for target setpoints that are far away, we linearly interpolate between the current and the target pose, such that the intermediate goal is only a limited distance away from the current pose.

\subsection{Fixed vs. Adaptive Virtual Stereo Baseline}
\label{ssec:fixed_vs_adaptive_baseline}
\begin{figure}[t]
    \centering
    \vspace{2mm}
    ~ 
    \begin{subfigure}[b]{0.99\columnwidth}
        \includegraphics[width=1.0\columnwidth]{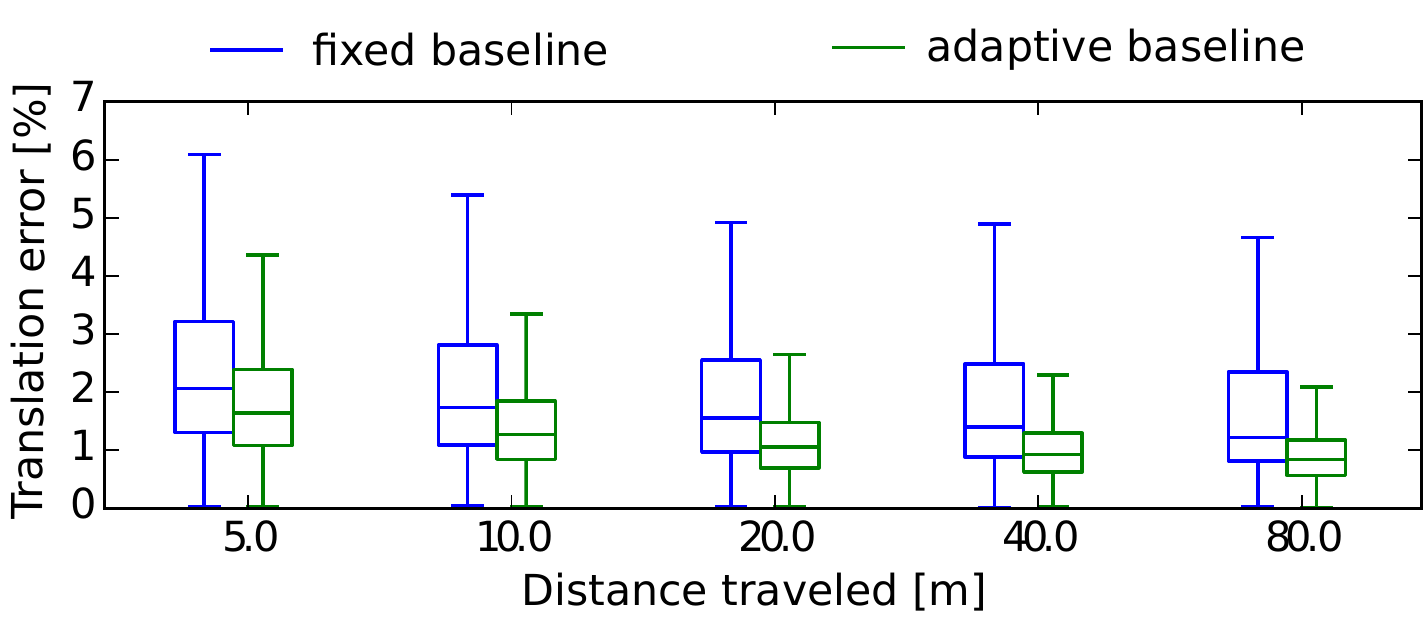}
        \caption{agent A}
        \label{fig:fix_vs_adaptive_rel_trans_A}
    \end{subfigure}
	\begin{subfigure}[b]{0.99\columnwidth}
        \includegraphics[width=1.0\columnwidth]{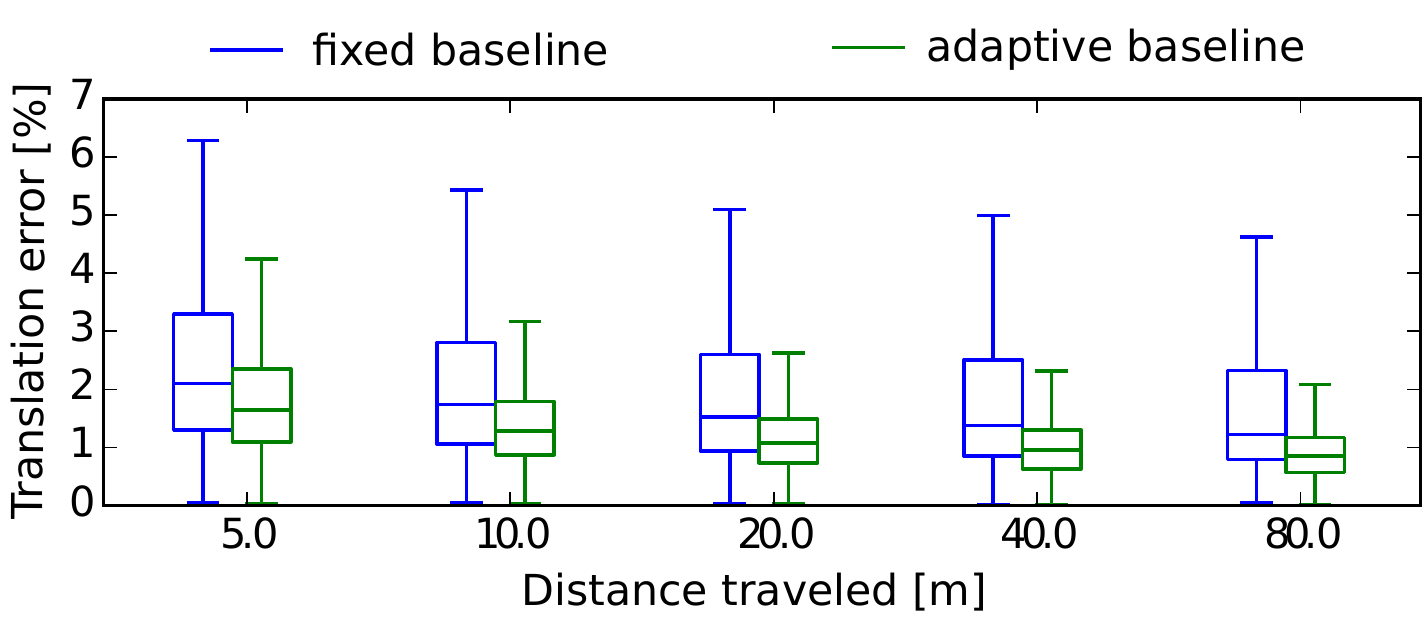}
        \caption{agent B}
        \label{fig:fix_vs_adaptive_rel_trans_B}
    \end{subfigure}
    \caption{Comparison of the relative translation error over all runs on all datasets. The results for agent A are shown in (a), whereas (b) shows the corresponding errors for agent B.}
\label{fig:fix_vs_adaptive_rel_trans_comp}
\end{figure}

\begin{table*}[t]
\centering
\begin{tabular}{l|c|c|c|c|c|c|c|c|c|c|c|c|}
\cline{2-13}
 &
  \multicolumn{6}{c|}{\textbf{Fixed Baseline}} &
  \multicolumn{6}{c|}{\textbf{Adaptive Baseline}} \\ \cline{2-13} 
 &
  \multicolumn{2}{c|}{\textbf{agent A}} &
  \multicolumn{2}{c|}{\textbf{agent B}} &
  \multicolumn{2}{c|}{\textbf{combined}} &
  \multicolumn{2}{c|}{\textbf{agent A}} &
  \multicolumn{2}{c|}{\textbf{agent B}} &
  \multicolumn{2}{c|}{\textbf{combined}} \\ \cline{2-13} 
 &
  \textbf{\begin{tabular}[c]{@{}c@{}}RMSE\\ {[}m{]}\end{tabular}} &
  \textbf{\begin{tabular}[c]{@{}c@{}}Scale\\ {[}\%{]}\end{tabular}} &
  \textbf{\begin{tabular}[c]{@{}c@{}}RMSE\\ {[}m{]}\end{tabular}} &
  \textbf{\begin{tabular}[c]{@{}c@{}}Scale\\ {[}\%{]}\end{tabular}} &
  \textbf{\begin{tabular}[c]{@{}c@{}}RMSE\\ {[}m{]}\end{tabular}} &
  \textbf{\begin{tabular}[c]{@{}c@{}}Scale\\ {[}\%{]}\end{tabular}} &
  \textbf{\begin{tabular}[c]{@{}c@{}}RMSE\\ {[}m{]}\end{tabular}} &
  \textbf{\begin{tabular}[c]{@{}c@{}}Scale\\ {[}\%{]}\end{tabular}} &
  \textbf{\begin{tabular}[c]{@{}c@{}}RMSE\\ {[}m{]}\end{tabular}} &
  \textbf{\begin{tabular}[c]{@{}c@{}}Scale\\ {[}\%{]}\end{tabular}} &
  \textbf{\begin{tabular}[c]{@{}c@{}}RMSE\\ {[}m{]}\end{tabular}} &
  \textbf{\begin{tabular}[c]{@{}c@{}}Scale\\ {[}\%{]}\end{tabular}} \\ \hline
\multicolumn{1}{|l|}{\textbf{Scene 1 - 1}} &
  3.73 & 2.28 & 3.68 &  2.31 &  3.70 & 2.29 & 2.16 & 0.66 & 2.10 & 0.66 & 2.13 & 0.66 \\ \hline
\multicolumn{1}{|l|}{\textbf{Scene 2 - 1}} &
  0.92 & 0.39 & 0.92 & 0.39 & 0.92 & 0.39 & 0.81 & 0.14 & 0.82 & 0.15 & 0.81 & 0.14 \\ \hline
\multicolumn{1}{|l|}{\textbf{Scene 3 - 1}} &
  1.54 & 1.19 & 1.51 & 1.16 & 1.52 & 1.17 & 0.48 & 0.33 & 0.48 & 0.32 & 0.48 & 0.32 \\ \hline
\multicolumn{1}{|l|}{\textbf{Scene 3 - 2}} &
  1.79 & 1.63 & 1.79 & 1.66 & 1.79 & 1.64 & 0.77 & 0.24 & 0.74 & 0.25 & 0.76 & 0.24 \\ \hline
\multicolumn{1}{|l|}{\textbf{Scene 4 - 1}} &
  2.71 & 1.95 & 2.73 & 1.92 & 2.72 & 1.93 & 0.88 & 0.67 & 0.89 & 0.71 & 0.89 & 0.69 \\ \hline
\end{tabular}
\caption{Comparison of the absolute trajectory errors when using a fixed baseline versus using the adaptive baseline control scheme. All the the reported values are obtained as average over 3 runs. To illustrate the consistency of the distributed approach, the errors of the individual agents (aligned individually) as well as the error of the combined trajectory are reported.}
\label{tab:comparison_fix_vs_variable_baseline}
\end{table*}

In this section, we evaluate the influence of actively adapting the virtual stereo baseline, i.e. a fixed $\alpha_{t}$, between the two \glspl{uav} against maintaining a fixed target distance between the two aircraft.
As this generally cannot be achieved using the exact same trajectories, for every dataset, we create two versions; one using a fixed baseline of $2$m and another, where the baseline gets constantly adjusted to achieve a triangulation angle $\alpha_{t}$ (of $10^{\circ}$).
However, the waypoints as well as the simulation parameters are otherwise chosen to be identical.
The estimator parameters in both versions are identical.
In particular, the \gls{kf} interval is set to $0.15$ seconds and the trajectory horizon to $5.0$ seconds for all the experiments in this section.
In total, five datasets are generated on the four scenes shown in \reffig{fig:photorealistic_models}, where the waypoints are chosen to follow mostly exploratory paths with some height variation, leading to scene depths ranging between $~25-120$m. 
The resulting error statistics of the obtained odometry estimates for both cases are summarized in \reffig{fig:fix_vs_adaptive_rel_trans_comp}.
As it can be observed, the position drift is significantly reduced when adapting the baseline versus having a fixed baseline.
The relative translation error of the fixed baseline approach ranges between $1.8-2.5\%$, while using an adaptive baseline the error can be reduced to $0.9-1.9\%$, corresponding to up to a twofold reduction. \\
Comparing the global \gls{rmse} in the agents' trajectories on the datasets presented in \reftab{tab:comparison_fix_vs_variable_baseline}, the differences are clearly visible and are, at times, up to a factor of 3.
However, as it can be seen in \reftab{tab:comparison_fix_vs_variable_baseline}, the fixed baseline approach mainly suffers from worse scale estimates, which in return, leads to increased \glspl{rmse} on the trajectory.
This is not surprising and supports our thesis that the correct scale estimation becomes a crucial element for robust and stable pose estimation at high altitudes. 
For the generated datasets used here, the fixed stereo baseline of $2$m, is rather small compared to the scenes' depth, which are to a large extent, given by the flight altitude, and therefore, the scale estimation becomes more uncertain than for larger baselines.
On the other hand, on the dataset of Scene 2, no significant difference between the adaptive and the fixed baseline approach can be observed, which can be explained by the fact, that the corresponding scene has more depth variations to it, which leads to having some close (e.g. side walls), but also some farther away parts of the scene (e.g. ground) in the view overlap, which results in smaller differences between the adaptive baseline and the fixed one. \\
Besides the advantage of actively controlling the baseline between the agents, also the consistency between the estimates of the two agents is indicated in \reftab{tab:comparison_fix_vs_variable_baseline}.
To illustrate this, we report the global \gls{rmse} of the individual agents aligning their trajectories independently, as well as the resulting errors when aligning both trajectories treating the two trajectories as a single one.
If the two trajectories were inconsistent with respect to each other, an increase in the combined trajectories error should be observed, however, throughout all datasets this is not the case, indicating the consistency of the distributed estimate.
Note that there is no noticeable difference in the consistency between the adaptive and the fixed baseline approach, which indicates that the distributed approach works reliably even when the estimate itself becomes more uncertain (i.e. fixed baseline).

\section{Conclusion}\label{sec:conclusions}
In this article, we present a novel framework using two \glspl{uav}, equipped with one \gls{imu} and one monocular camera each, while measuring the relative distance between them using an Ultrawideband module in order to compute the 6DoF pose estimation for both \glspl{uav} in real-time to estimate the scene in a virtual-stereo setup.
The pipeline is implemented in a decentralized fashion allowing each agent to hold its own estimate of the map, enabling low-latency pose estimation that does not depend on network delays. 
In order to ensure consistency across the agents, a consensus based optimization scheme is employed.
Using the two agents as a virtual stereo camera with adjustable baseline, potentially large scene depths can be handled, which are problematic for existing \gls{vio} system using monocular or (fixed-baseline) stereo rigs. \\
A thorough experimental analysis using synthetic, photorealistic data reveals the ability of the proposed approach to reliably estimate the pose of the agents even at high altitudes.
Using trajectories with increasing altitudes, we show the problematic behavior of existing the stereo \gls{vio} methods at high altitudes, while the proposed approach is able to maintain high quality estimation of both \gls{uav} agents' poses.
Furthermore, the comparison with state-the-art stereo inertial methods demonstrates, that the proposed approach already proves advantageous in terms of estimation accuracy at altitudes marginally higher than $15$m.
Employing a simple formation controller, which adjusts the baseline between the two agents depending on the observed scene depth, the benefit of having the ability to adjust the stereo baseline on-the-fly is be demonstrated, achieving a nearly twofold reduction in the pose estimation error compared to a fixed target baseline. \\
The applicability of the approach is verified by reporting practical timing and bandwidth measurements.
Owing to the decentralized approach, the proposed system achieves an average latency of $11$ms for the pose tracking.
The required bandwidth of the overall system remains under $250$KB/s and therefore, can easily be handled by a standard WiFi module.\\
Future work includes employing the system in the field and perform extensive outdoor experiments.
Furthermore, it would be interesting to investigate non-uniform \gls{kf} intervals, to dynamically adjust them depending on the performed motions.
Also the investigation of the proposed system to be used as virtual stereo-camera for 3D reconstruction would be highly interesting as this potentially allows to quickly obtain a coarse reconstruction of a large areas. \\
Following the demonstration of the ability of the proposed method to drastically increase the fidelity of estimates at high flying altitudes using two UAV agents, future work will aim to leverage the power of the asynchronous estimation framework proposed to scale up to bigger number of agents in the air. 
In this way, we aim to push towards tightly collaborating multi-agent SLAM in a distributed architecture.

%
\IEEEpeerreviewmaketitle


%

\appendices
\section{}\label{sec:appendix_1}
\subsection{Rotation Calculus}
\label{ssec:rotation_calculus}
First we introduce the exponential map, mapping a vector $\pmb{\varphi} \in \mathbb{R}^{3}$ to a rotation $\mathbf{q}$
\begin{equation}
\mathbf{q} = \mathrm{exp} \left(\pmb{\varphi} \right) = \left( q_{0}, \breve{\mathbf{q}} \right) = \left(\cos(\lVert \pmb{\varphi} \rVert / 2), \sin(\lVert \pmb{\varphi} \rVert / 2) \frac{\pmb{\varphi}}{\lVert \pmb{\varphi} \rVert} \right) ~,
\end{equation}
where $q_{0}$ is the real-part and $\breve{\mathbf{q}}$ is the imaginary part of the quaternion $\mathbf{q}$.
The inverse mapping, the logarithmic map, maps a quaternion $\mathbf{q}$ to its corresponding tangent vector $\pmb{\varphi}$:
\begin{equation}
\pmb{\varphi} = \mathrm{log}(\mathbf{q}) = 2 \mathrm{atan2}\left( \lVert \breve{\mathbf{q}} \rVert, q_{0} \right)\frac{\breve{\mathbf{q}}}{\lVert \breve{\mathbf{q}} \rVert} ~.
\end{equation}
Using these mappings, boxplus and boxminus operations adopting the functions of addition and subtraction \cite{Hertzberg:etal:INFUS2011} can be constructed as follows:
\begin{align}
\boxplus &: SO(3) \times \mathbb{R}^{3} \rightarrow SO(3), \\
 &\mathbf{q}, \pmb{\varphi} \mapsto \mathbf{q} \circ \mathrm{exp}(\pmb{\varphi}) \nonumber \\
 \boxminus &: SO(3) \times SO(3) \rightarrow \mathbb{R}^{3} \\
 &\mathbf{q}_{1}, \mathbf{q}_{2} \mapsto \mathrm{log}(\mathbf{q}_{2}^{-1} \circ \mathbf{q}_{1}) \nonumber ~,
\end{align}
where $\circ$ indicates the concatenation of two quaternions.

\subsection{Pose Tracking and Map Point EKFs}
\label{ssec:jacobians_pose_tracking_ekf}
The state transition jacobian used in Eq. \eqref{eq:ekf_covariance_prediction} is given by:
\begin{equation}
\resizebox{0.99\columnwidth}{!}{$
\mathbf{F} = \begin{bmatrix}
\mathbf{I}_{2 \times 2}  & \mathbf{0} & \mathbf{0} & \mathbf{0} & \mathbf{0} & \mathbf{0} \\
 & \frac{\partial \hat{\mathbf{q}}_{MS}^{k+1}}{\partial \mathbf{q}_{MS}^{k}} & \mathbf{0} & \mathbf{0} & \mathbf{0} & \frac{\partial \hat{\mathbf{q}}_{MS}^{k+1}}{\partial \mathbf{b}_{g}^{k}} \\
 \frac{\partial \prescript{}{S}{\hat{\mathbf{v}}}^{k+1}}{\partial \mathbf{q}_{WM}^{k}} & \frac{\partial \prescript{}{S}{\hat{\mathbf{v}}}^{k+1}}{\mathbf{q}_{MS}^{k}} & \mathbf{0} & \frac{\partial \prescript{}{S}{\hat{\mathbf{v}}}^{k+1}}{\prescript{}{S}{\mathbf{v}}^{k}} & -\Delta t \mathbf{I}_{3\times 3} & \Delta t \left[ \prescript{}{S}{\mathbf{v}}^{k} \right]^{\times} \\
\mathbf{0} & \mathbf{0} & \mathbf{0} & \mathbf{0} & \mathbf{I}_{3\times 3} & \\
\mathbf{0} & \mathbf{0} & \mathbf{0} & \mathbf{0} & \mathbf{0} & \mathbf{I}_{3\times 3} 
\end{bmatrix}
$}
\end{equation}
with
\begin{align}
\frac{\partial \hat{\mathbf{q}}_{MS}^{k+1}}{\partial \mathbf{q}_{MS}^{k}} &= \mathbf{R}(\mathrm{exp}(\Delta t \prescript{}{S}{\pmb{\omega}}_{WS}^{k}))^{T} \\
\frac{\partial \hat{\mathbf{q}}_{MS}^{k+1}}{\partial \mathbf{b}_{g}^{k}} &=  -\Delta t \Gamma(\Delta t \prescript{}{S}{\pmb{\omega}}_{WS}) \\
\frac{\partial \prescript{}{S}{\hat{\mathbf{v}}}^{k+1}}{\partial \mathbf{q}_{WM}^{k}} &= \Delta t \mathbf{R}(\mathbf{q}_{MS}^{k})^{T} \left[ \mathbf{R}(\mathbf{q}_{WM}^{k})^{T} \mathbf{g}  \right]^{\times} \mathbf{J}_{rp}(\mathbf{q}_{WM}^{k}) \\
\frac{\partial \prescript{}{S}{\hat{\mathbf{v}}}^{k+1}}{\mathbf{q}_{MS}^{k}} &= \Delta t \left[ \mathbf{R}(\mathbf{q}_{MS}^{-1} \circ \mathbf{q}_{WM}^{-1}) \mathbf{g} \right]^{\times} \\
\frac{\partial \prescript{}{S}{\hat{\mathbf{v}}}^{k+1}}{\prescript{}{S}{\mathbf{v}}^{k}} &= \mathbf{I}_{3\times 3} - \Delta t \left[ \prescript{}{S}{\pmb{\omega}}_{WS} \right]^{\times} ~,
\end{align}
where $\Gamma(\cdot)$ is the jacobian of the exponential map given by
\begin{equation}
\Gamma(\pmb{\varphi}) = \mathbf{I}_{3\times 3} -  \frac{1 - \cos(\lVert \pmb{\varphi} \rVert)}{\lVert \pmb{\varphi} \rVert^{2}} \left[ \pmb{\varphi} \right]^{\times} + \frac{\lVert \pmb{\varphi} \rVert - \sin(\lVert \pmb{\varphi} \rVert)}{\lVert \pmb{\varphi} \rVert^{3}} (\left[ \pmb{\varphi} \right]^{\times})^{2}
\end{equation}
and $\mathbf{J}_{rp}(\mathbf{q})$ denotes the jacobian of the local roll-pitch parameterization and is given by
\begin{equation}
\mathbf{J}_{rp}(\mathbf{q}) = \begin{bmatrix}
1 & 0 & 0 \\
0 & \cos(roll(\mathbf{q})) & -\sin(roll(\mathbf{q}))
\end{bmatrix}^{T} ~.
\end{equation}
The function $roll(\mathbf{q})$ extracts the roll angle from a quaternion $\mathbf{q}$. \\
Analogously, the jacobian matrix of the prediction noise can be described as
\begin{equation}
\resizebox{0.99\columnwidth}{!}{$
\mathbf{G} = \begin{bmatrix}
\Delta t \mathbf{I}_{2 \times 2} & \mathbf{0} & \mathbf{0} & \mathbf{0} & \mathbf{0} & \mathbf{0} \\
\mathbf{0} & \mathbf{0} & \mathbf{0} &  \Delta t \Gamma(\prescript{}{S}{\pmb{\omega}}_{WS}) & \mathbf{0} & \mathbf{0} \\
\mathbf{0} & \Delta t \mathbf{R}(\mathbf{q}_{MS}^{k}) & \mathbf{0} & \mathbf{0} & \mathbf{0} & \mathbf{0} \\
\mathbf{0} & \mathbf{0} & \Delta t \mathbf{I}_{3\times 3} & \Delta t \left[ \prescript{}{S}{\mathbf{v}}^{k} \right]^{\times} & \mathbf{0} & \mathbf{0} \\
\mathbf{0} & \mathbf{0} & \mathbf{0} & \mathbf{0} & \Delta t \mathbf{I}_{3\times 3} & \mathbf{0} \\
\mathbf{0} & \mathbf{0} & \mathbf{0} & \mathbf{0} & \mathbf{0} & \Delta t \mathbf{I}_{3\times3}
\end{bmatrix}
$}
\end{equation}
The corresponding noise matrix $\mathbf{W}$ can be written as
\begin{equation}
\resizebox{0.99\columnwidth}{!}{$
\mathbf{W} = \begin{bmatrix}
\sigma_{g}^{2} \mathbf{I}_{2 \times 2} & \mathbf{0} & \mathbf{0} & \mathbf{0} & \mathbf{0} & \mathbf{0} \\
\mathbf{0} & \sigma_{v}^{2} \mathbf{I}_{3 \times 3} & \mathbf{0} & \mathbf{0} & \mathbf{0} & \mathbf{0} \\
\mathbf{0} & \mathbf{0} & \sigma_{a}^{2} \mathbf{I}_{3 \times 3} & \mathbf{0} & \mathbf{0} & \mathbf{0} \\
\mathbf{0} & \mathbf{0} & \mathbf{0} & \sigma_{\omega}^{2} \mathbf{I}_{3 \times 3} & \mathbf{0} & \mathbf{0}  \\
\mathbf{0} & \mathbf{0} & \mathbf{0} & \mathbf{0} & \sigma_{b_{a}}^{2} \mathbf{I}_{3\times 3} & \mathbf{0} \\
\mathbf{0} & \mathbf{0} & \mathbf{0} & \mathbf{0} & \mathbf{0} & \sigma_{b_{g}}^{2} \mathbf{I}_{3 \times 3} 
\end{bmatrix}
$} ~,
\end{equation}
where $\sigma_{i}$ represents the discrete time noise of the corresponding variable $i$. \\
The measurement jacobian as used in Eq. \eqref{eq:ekf_innovation_covariance} for a single observation of the \gls{mp} $i$ is given by:
\begin{equation}
\mathbf{H}_{i} = \begin{bmatrix} \mathbf{0}_{2 \times 2} & \frac{\partial \pi \left(\prescript{}{C}{\mathbf{m}}_{i} \right) }{\partial \hat{\mathbf{q}}_{MS}^{k+1}} & \frac{\partial \pi \left(\prescript{}{C}{\mathbf{m}}_{i} \right) }{\partial \hat{\mathbf{p}}_{MS}^{k+1}} & \mathbf{0}_{2 \times 9}
\end{bmatrix} ~,
\end{equation}
where 
\begin{align}
\frac{\partial \pi \left(\prescript{}{C}{\mathbf{m}}_{i} \right) }{\partial \hat{\mathbf{q}}_{MS}^{k+1}} &= -\frac{\partial \pi \left( \prescript{}{C}{\mathbf{m}}_{i}\right)}{\partial \prescript{}{C}{\mathbf{m}}_{i}} \mathbf{R}(\mathbf{q}_{CS})\left[ \prescript{}{S}{\mathbf{m}}_{i} \right]^{\times} \\
\frac{\partial \pi \left(\prescript{}{C}{\mathbf{m}}_{i} \right) }{\partial \hat{\mathbf{p}}_{MS}^{k+1}} &= \frac{\partial \pi \left( \prescript{}{C}{\mathbf{m}}_{i}\right)}{\partial \prescript{}{C}{\mathbf{m}}_{i}} \mathbf{R}(\mathbf{q}_{CS}) \mathbf{R}(\hat{\mathbf{q}}_{MS}^{k+1})^{T} ~,
\end{align}
where $\frac{\partial \pi \left( \prescript{}{C}{\mathbf{m}}_{i}\right)}{\partial \prescript{}{C}{\mathbf{m}}_{i}}$ is the jacobian of the camera model including the lens distortion. \\


The jacobian $\mathbf{J}_{i,j}$ used to compute the \gls{mp} uncertainty in Eq. \eqref{eq:initial_map_point_uncertainty}, is given by
\begin{equation}
\mathbf{J}_{i,j} = \begin{bmatrix}
\frac{\partial \pi \left( \prescript{}{C_{i}}{\mathbf{m}}\right)}{\partial \prescript{}{C_{i}}{\mathbf{m}}} \frac{\partial \prescript{}{C_{i}}{\mathbf{m}}}{\partial \prescript{}{M}{\mathbf{m}}} & \frac{\partial \pi \left( \prescript{}{C_{i}}{\mathbf{m}}\right)}{\partial \prescript{}{C_{i}}{\mathbf{m}}} \frac{\partial \prescript{}{C_{i}}{\mathbf{m}}}{\partial \mathbf{T}_{MS_{i}}} & \mathbf{0}_{2 \times 6} \\
\frac{\partial \pi \left( \prescript{}{C_{j}}{\mathbf{m}}\right)}{\partial \prescript{}{C_{j}}{\mathbf{m}}} \frac{\partial \prescript{}{C_{j}}{\mathbf{m}}}{\partial \prescript{}{M}{\mathbf{m}}} & \mathbf{0}_{2 \times 6} & \frac{\partial \pi \left( \prescript{}{C_{j}}{\mathbf{m}}\right)}{\partial \prescript{}{C_{j}}{\mathbf{m}}} \frac{\partial \prescript{}{C_{j}}{\mathbf{m}}}{\partial \mathbf{T}_{MS_{j}}} \\
\mathbf{0}_{6 \times 3} & \mathbf{I}_{6 \times 6} & \mathbf{0}_{6 \times 6} \\
\mathbf{0}_{6 \times 3} & \mathbf{0}_{6 \times 6} & \mathbf{I}_{6 \times 6}
\end{bmatrix} ~,
\end{equation}
where again the $\frac{\partial \pi \left( \prescript{}{C}{\mathbf{m}}\right)}{\partial \prescript{}{C}{\mathbf{m}}}$ corresponds to the projection model.
The remaining terms are given by
\begin{align}
\frac{\partial \prescript{}{C_{k}}{\mathbf{m}}}{\partial \prescript{}{M}{\mathbf{m}}} &= \mathbf{R}_{CS} \mathbf{R}_{MS_{k}}^{T} \\
\frac{\partial \prescript{}{C_{k}}{\mathbf{m}}}{\partial \mathbf{T}_{MS_{k}}} &= \begin{bmatrix}
\left[ \mathbf{R}_{MC}^{T} (\prescript{}{M}{\mathbf{m}} - \mathbf{p}_{MC}) \right]^{\times} & \mathbf{0}_{3\times 3} \\
\mathbf{0}_{3 \times 3} & -\mathbf{R}_{MC}^{T}
\end{bmatrix} ~,
\end{align}
where $k \in [i, j]$.


\ifCLASSOPTIONcaptionsoff
  \newpage
\fi



%
{\small
\bibliographystyle{ieee}
\bibliography{./bibliography/references}
}





\end{document}